%% file: main.tex
\documentclass[journal,twoside,web]{ieeecolor}
\usepackage{jsen}
\usepackage{cite}
\usepackage{amsmath,amssymb,amsfonts}
\usepackage{algorithmic}
\usepackage{graphicx}
\usepackage{subcaption}
\usepackage{textcomp}
\usepackage{wrapfig}
\usepackage[colorinlistoftodos]{todonotes}

\def\BibTeX{{\rm B\kern-.05em{\sc i\kern-.025em b}\kern-.08em
    T\kern-.1667em\lower.7ex\hbox{E}\kern-.125emX}}
\definecolor{abstractbg}{rgb}{0.89804,0.94510,0.83137}
\begin{document}
\title{PATH: Continuous Target Sensing among Autonomous Cooperative Drones}
\author{
Heegyeong Kim,
Alice James,
Avishkar Seth,
Endrowednes Kuantama,
Jane Williamson,
Yimeng Feng, 
Richard Han
\thanks{
Heegyeong Kim, Alice James, Avishkar Seth, Endrowednes Kuantama,
Yimeng Feng, and Richard Han are with the School of Computing,
Macquarie University, Sydney, NSW, Australia
(e-mail: heegyeong.kim@mq.edu.au; alice.james@mq.edu.au;
avishkar.seth@mq.edu.au; endrowednes.kuantama@mq.edu.au;
yimeng.feng@mq.edu.au; richard.han@mq.edu.au).
}
\thanks{
Jane Williamson is with the School of Natural Sciences,
Macquarie University, Sydney, NSW, Australia.
}
}

\IEEEtitleabstractindextext{%
\fcolorbox{abstractbg}{abstractbg}{%
\begin{minipage}{\textwidth}%
\begin{wrapfigure}[15]{r}{2.2in}%
\includegraphics[width=2.1in]{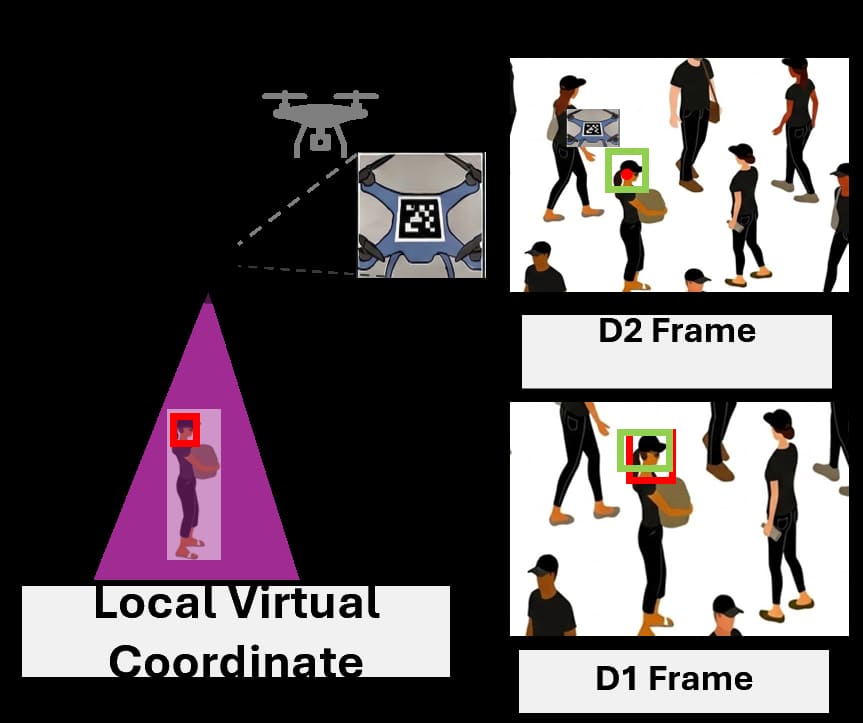}%
\label{fig:overall}
\end{wrapfigure}%
\begin{abstract}
Continuous target sensing by uncrewed aerial vehicles (UAVs) is constrained by limited flight endurance, motivating the transfer of tracking responsibility between cooperating UAVs. Such a handoff requires the receiver to identify the same physical target currently tracked by the sender despite differences in viewpoint, scale, and target appearance. Existing approaches based on global target localization or appearance-based cross-view association are limited by positioning uncertainty or ambiguous visual features. This paper presents Perspective Alignment \& Tracking Handoff (\textbf{PATH}), a platform-agnostic, geometry-assisted sensing and verification framework for target handoff between two moving UAVs. The sender reconstructs the tracked target as a metric 3D point using RGB-D sensing, while the receiver estimates its relative pose from a fiducial observation and projects the transmitted target point into its own image as a spatial prior for target acquisition. The receiver-generated candidate is then returned to the sender and verified through a cross-view Mutual Agreement Handshake before tracking responsibility is transferred. Real-world UAV experiments show mean relative-position and target-position errors of 0.047~m and 0.030~m, respectively. Under visually ambiguous conditions, PATH achieves 96.0\% frame-level receiver-side target acquisition accuracy, with 2.0\% false-positive and 2.0\% false-negative rates. A sensor-error sensitivity analysis shows that relative-pose uncertainty is the dominant contributor to receiver-view projection error. The implementation operates at video rate with compact inter-UAV communication below 16~kB/s at 60~Hz, demonstrating the feasibility of lightweight geometry-assisted target handoff on resource-constrained UAV platforms.
\end{abstract}

\end{minipage}}}

\maketitle

\input{Sections/Introduction}
\input{Sections/RelatedWork}
\input{Sections/System}
\input{Sections/Setup}
\input{Sections/Results}
\input{Sections/discussion_conclusion}

\bibliographystyle{IEEEtran}
\bibliography{ref}

\end{document}

%% file: Sections/Introduction.tex
\section{Introduction}
Uncrewed Aerial Vehicles (UAVs) equipped with onboard visual sensors have become widely used for dynamic target tracking in applications including wildlife monitoring, security surveillance~\cite{mishra2020drone,fang2024strategies,thakur2021artificial}, environmental observation~\cite{akram2024dronessl, ASADZADEH2022109633}, event tracking~\cite{han2024event}, and search and rescue operations~\cite{yeom2024thermal,alsamhi2022uav,dahal2021design,lun2022target}. Recent sensor-based UAV tracking systems have further demonstrated real-time target detection, state estimation, and tracking using onboard sensing~\cite{10552170} and edge computation~\cite{10778212,visiondronedetect2025wang}
For long duration missions, however, the limited endurance of a single UAV motivates cooperative operation in which a second UAV assumes tracking responsibility from the current platform ~\cite{LiangSrigrarom_ICUAS2021, Zhan_Chen_Chen_Zhang_2025, bauer2024persistent}. Such a handoff requires more than detecting an object in the receiver view: the receiver UAV must acquire the same physical target that is currently being tracked by the sender UAV.

\begin{figure}
    \centering
    \includegraphics[width=0.8\linewidth]{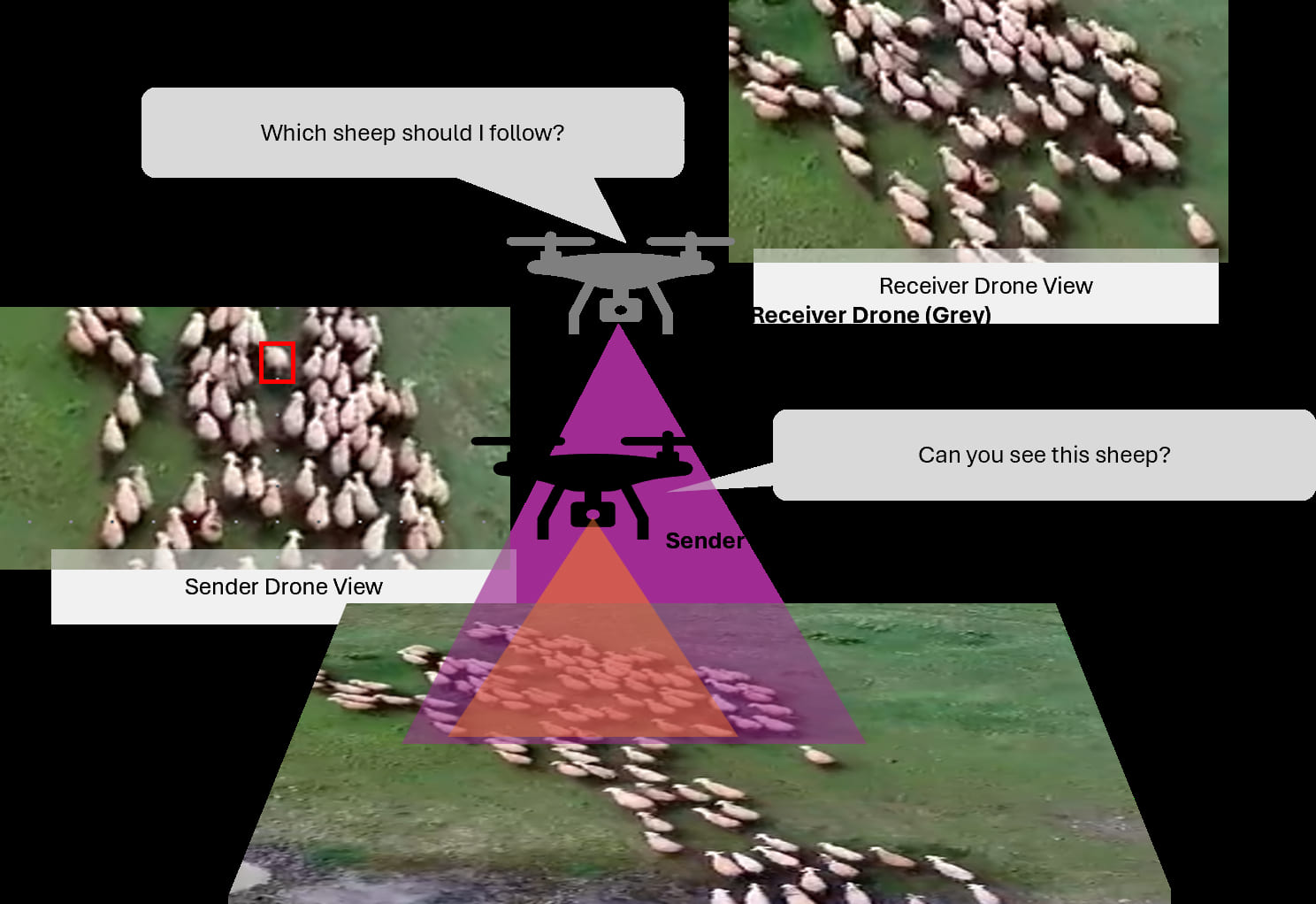}
    \caption{The handoff challenge: Continuous tracking.}
    \label{fig:overall}
\end{figure}

{Fig.~\ref{fig:overall} illustrates a cross view sensing problem between two independently moving cameras. During handoff sender UAV ($D_1$) must convey precisely which physical target is being tracked so that receiver UAV ($D_2$) can identify the same target from its own viewpoint. This is challenging because the two UAVs observe the scene from different, time varying positions and orientations, and $D_2$ may simultaneously observe multiple visually similar candidate objects in its field of view (FoV). A detector or tracker running independently on $D_2$ can identify candidate objects, but does not by itself determine which candidate, if any, corresponds to the target currently tracked by $D_1$. The handoff problem therefore requires a mechanism for establishing correspondence between observations made from the two moving viewpoints.


Existing approaches provide only partial solutions to this cross-view handoff problem. Global GNSS coordinates are not well suited to establishing target correspondence between two nearby UAVs because the handoff requires accurate relative geometry between the sender, receiver, and target rather than their independent positions in a global frame. Standard GNSS is vulnerable to multipath, signal blockage, and degraded satellite visibility~\cite{10620329}, while centimeter-level RTK positioning requires additional infrastructure~\cite{11271135}, correction communication, and sufficiently favorable operating conditions. Importantly, multipath remains a dominant error source even for short-baseline RTK configurations~\cite{10620329, 11271135}. These dependencies increase deployment complexity while still not directly resolving which of several closely spaced observations corresponds to the target tracked by the sender. PATH therefore avoids reliance on global positioning for handoff verification and instead constructs the required geometry directly in a local inter-UAV reference frame. Active optical designation using a laser could instead explicitly indicate the target of interest. However, such a system requires an additional active pointing and tracking mechanism, and laser illumination can interfere with optical sensing~\cite{kuantama2024laser}. Moreover, laser exposure can alter animal behavior, which limits its suitability for applications requiring passive wildlife~\cite{elbers2021efficacy,eisenbeiser2022gills,grigg2024associations}. Appearance-based feature matching or learned re-identification provides another alternative, but as we will show correspondence may degrade under viewpoint and scale changes, weak visual features, or visually similar distractors, while introducing an additional matching stage on onboard hardware.

We therefore present \textbf{PATH} (\textbf{P}erspective \textbf{A}lignment and \textbf{T}racking \textbf{H}andoff), a geometry-assisted sensing framework for UAV target handoff. PATH constructs a handoff-local 3D representation using calibrated onboard sensing and fiducial-based relative-pose estimation between the two UAVs. The target observed by $D_1$ is reconstructed as a 3D point in this local frame and transmitted to $D_2$, where it is projected into the receiver image and used as a spatial prior for local target acquisition. The target itself remains uninstrumented. Once $D_2$ acquires a candidate, its detector-generated bounding box is returned to $D_1$, transformed into the sender image frame, and compared with the active target observation through the Mutual Agreement Handshake to verify the correct target is being tracked before tracking responsibility is transferred.

PATH is intended as a lightweight, accurate, low cost, robust, cross-platform and real time mechanism for transferring target tracking from one UAV to another in a verified manner.  Relative pose is estimated locally by $D_2$, and the inter-UAV exchange is limited to compact geometric observations rather than raw images, video streams, depth frames, or visual feature descriptors. PATH complements the detector or tracker already running on each UAV rather than replaces it.  The present study focuses on handoff-local geometry, receiver-side target acquisition, and cross-view verification. 
We evaluate PATH on real UAV platforms through controlled and dynamic indoor experiments, visually ambiguous outdoor target-acquisition experiments, end-to-end handoff trials, and onboard computation and communication measurements. We also analyze how representative RGB-D and relative-pose sensing errors propagate to the receiver-view target projection.

The main contributions of this article are as follows:
\begin{itemize}

\item We formulate cooperative UAV target handoff as a cross-view sensing problem in which the receiver must acquire and verify the same physical target observed by the sender before tracking responsibility is transferred.

\item We develop PATH, 
a platform-agnostic geometric handoff framework that combines RGB-D target reconstruction, inter-UAV relative-pose sensing, and cross-view projection to transfer a target location between moving UAV views, followed by a Mutual Agreement Handshake for geometric verification of the returned candidate.

\item We implement and evaluate PATH on heterogeneous UAV platforms, including geometric accuracy, receiver-side target acquisition under visually ambiguous conditions, complete handoff reliability, sensor-error sensitivity, and onboard computation and communication cost.

\end{itemize}

%% file: Sections/RelatedWork.tex
\section{Related Work}
\label{rel_work}

\subsection{Target Localization and Active Target Indication}

A straightforward approach to cooperative target handoff is to provide the receiver UAV with the target position in a shared global reference frame. GNSS-based target localization can provide such a spatial cue~\cite{Upadhyay_Rawat_Deb_2021}. However, standalone GNSS accuracy cannot always be relied upon to distinguish closely spaced targets, since positioning can degrade or become unavailable in environments with obstructions such as canyons, buildings, and hills
~\cite{tavasci2024reliability}. PATH's visually based perspective alignment is self-contained and hence avoids such limitations.

Active target indication provides another possible solution. Laser-based systems can explicitly mark an object of interest ~\cite{ming2022laser,haalck2023cater}, but stable pointing from a moving platform introduces additional sensing and control requirements. Laser illumination may also disturb or pose risks to tracked animals, which limits its suitability for passive wildlife observation ~\cite{elbers2021efficacy,eisenbeiser2022gills,grigg2024associations}. PATH instead keeps the tracked target uninstrumented and uses passive onboard visual sensing to establish the handoff geometry.

\subsection{Cross-View Visual Association}

Visual feature matching and learned re-identification provide a passive alternative for associating observations across cameras ~\cite{pengnianmultitrack2024_drones}. Re-identification and multi-camera tracking have been extensively studied using appearance, motion, and spatio-temporal information ~\cite{he2020city,hsu2021multi,liu2021city,yang2022box, specker2022improving,zhu2022visdrone,ye2022survey, LiangSrigrarom_ICUAS2021}. These approaches are effective when sufficient appearance information is available, but cross-view association becomes more difficult when targets have similar appearance or are observed under substantially different viewpoints and scales~\cite{kim2025continuous}.

Mobile-camera systems introduce an additional challenge because the relative geometry between cameras changes over time. Collaborative UAV systems have therefore explored the exchange of regions of interest and visual information for multi-platform detection and tracking ~\cite{Upadhyay_Rawat_Deb_2021}. Such visual association remains complementary to the approach considered here: rather than matching target appearance across the two UAV views, PATH uses the relative camera geometry to constrain where the sender-observed target should appear in the receiver image. The receiver detector is then used only to generate the candidate observation required for handoff verification.

\subsection{Cooperative Relative Localization and Geometric Sensing}

Cooperative localization methods combine measurements such as range, inertial sensing, vision, and inter-agent communication to estimate relative or global states~\cite{fu2017multi,lee2013cooperative}. Multi-sensor UAV systems similarly fuse RGB, thermal, inertial, or position measurements for target localization and search tasks ~\cite{cui2015drones}. These methods primarily address localization or shared-state estimation rather than verification that a receiver-side detection corresponds to the specific target currently tracked by another moving UAV.

PATH addresses this complementary handoff problem by constructing a local geometric representation between the sender and receiver at the time of transfer. A fiducial observation provides the inter-UAV relative pose, while RGB-D sensing reconstructs the sender-side target as a metric 3D point. This point is projected into the receiver image to provide a spatial prior for target acquisition, and the receiver's candidate is subsequently returned to the sender for geometric agreement before tracking responsibility is transferred. In this way, PATH combines relative geometric sensing and cross-view verification without requiring a globally calibrated multi-camera system or appearance-based inter-UAV feature matching.

%% file: Sections/system.tex
\section{Proposed PATH Verification System}

PATH provides a self verification mechanism for autonomous target handoff between two UAVs. The overall procedure consists of five steps. First, $D_2$ observes $D_1$ and establishes the shared loval coordinate frame. Second, $D_1$ sends the centroid of target bounding box to $D_2$. Third, $D_2$ reprojects this target location into its own image frame and detects the corresponding target. Fourth, $D_2$ sends the detected target bounding box back to $D_1$. Finally, $D_1$ evaluates the bounding boxes between the projected target prior and $D_2$'s detection using IoU, completing the handoff verification 

\subsection{Design Objectives and System Choices}
\label{sec:design_objectives}

PATH is designed as a cooperative UAV-to-UAV verification layer for target handoff. Rather than replacing the detector or tracker used by either UAV, it determines whether the receiver UAV ($D_2$) observes the same target that is being tracked by the sender UAV ($D_1$) before tracking responsibility is transferred. The formulation is not tied to a particular target category, provided that each UAV can produce a bounding box observation of the target. 

The system is designed for real-time, low-cost operation using non-invasive onboard sensing without instrumenting the tracked target. PATH therefore uses calibrated visual sensing together with a fiducial marker mounted only on $D_1$ to establish the relative geometry required for cross view projection. Existing detectors or trackers need only provide bounding boxes, while PATH performs geometric consistency verification without requiring feature level appearance matching. Cross view verification is performed locally using compact inter-UAV messages rather than raw image or video transmission or centralized visual association. The region-based verification is also designed to tolerate moderate sensing and bounding box variations during handoff. 
The handoff local geometric representation and verification procedure are described in the following subsections. 

\subsection{Handoff Geometric Representation}
\label{virtual_3D}

\begin{figure}[!h]
\centering
    \includegraphics[width=0.9\linewidth]{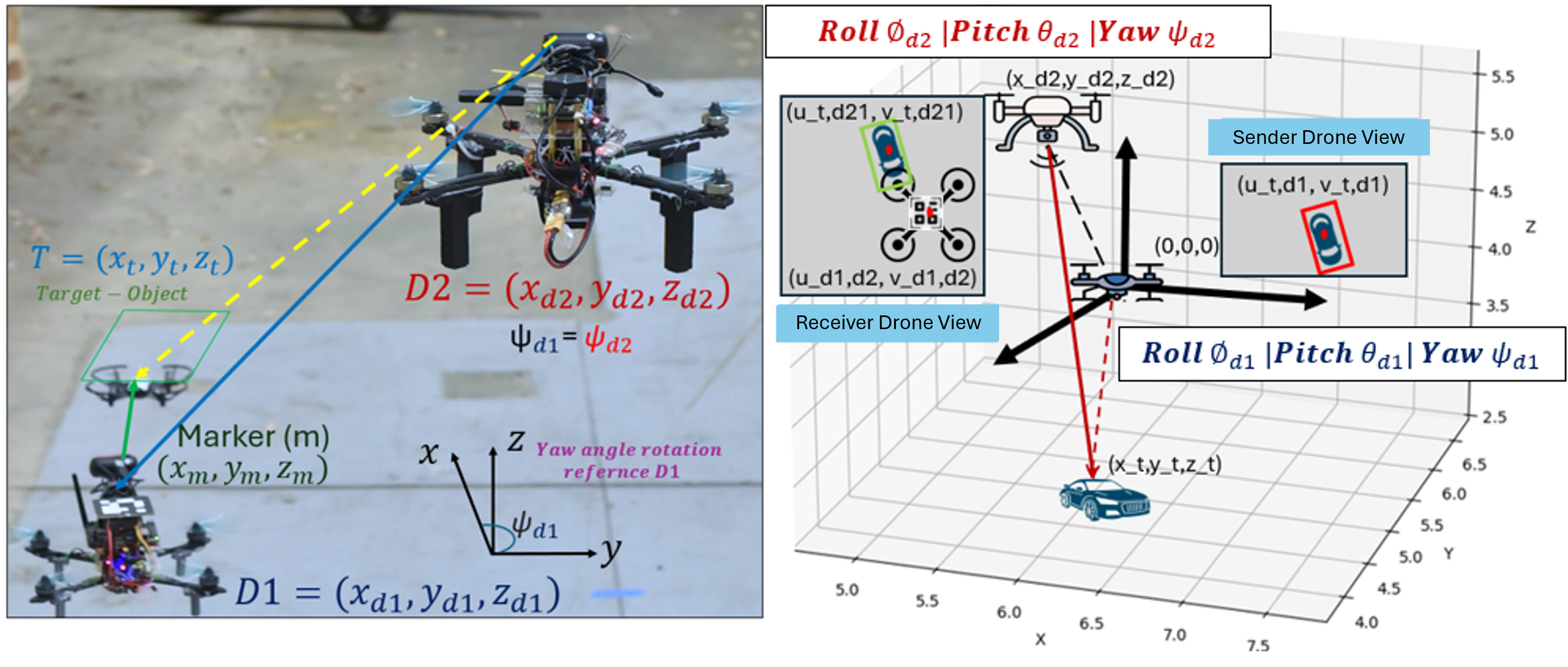}
    \caption{Local virtual coordinate between $D_1$ and $D_2$ to track the target (T).}
    \label{fig:coordinate}
\end{figure}

During handoff, PATH establishes a local coordinate frame $\mathcal{F}_v$ anchored to the sender UAV $D_1$. The origin is defined at the center of the fiducial marker rigidly mounted on $D_1$, which provides a fixed reference with respect to the $D_1$ body frame. The coordinate axes follow the orientation illustrated in Fig.~\ref{fig:coordinate}. This handoff local representation allows the target observation maintained by $D_1$ and the relative pose of the receiver UAV $D_2$ to be expressed with respect to the same spatial reference. 

When $D_2$ reaches the handoff region, it observes the fiducial marker rigidly mounted on $D_1$ and estimates its relative pose with respect to $\mathcal{F}_v$ ~\cite{803809}. Let ${}^{v}\mathbf{T}_{2}$ denote the homogeneous transformation from the receiver $D_2$ camera with respect to the handoff-local frame $\mathcal{F}_v$:

\begin{equation}
{}^{v}\mathbf{T}_{2}=
\begin{bmatrix}
\mathbf{R}_{v2} & \mathbf{t}_{v2}\\
\mathbf{0}^{T} & 1
\end{bmatrix}
\label{eq:Tv2}
\end{equation}

where $\mathbf{t}_{v2}$ represents the position of $D_2$ in $\mathcal{F}_v$ and $\mathbf{R}_{v2}$ represents its relative orientation. To project a point expressed in $\mathcal{F}_v$ into the $D_2$ camera frame, the inverse transformation is used:

\begin{equation}
{}^{2}\mathbf{T}_{v} = \left({}^{v}\mathbf{T}_{2}\right)^{-1}
\label{eq:T2v}
\end{equation}

At $D_1$, the center of the currently tracked target bounding box is denoted by \begin{equation}
\mathbf{p}_{t,1}=
\begin{bmatrix} 
u_{t,1} & v_{t,1} & 1
\end{bmatrix}^{T}
\end{equation}

The corresponding metric depth $Z_{t,1}$ is obtained from the registered depth image. Because lens distortion is retained in the sensing model, the measured pixel is first corrected using the calibrated intrinsic matrix $\mathbf{K}_1$ and distortion coefficients $\mathbf{D}_1=[k_1,k_2,p_1,p_2,k_3]$. Let $(x_{t,1}^{u},y_{t,1}^{u})$ denote the resulting undistorted normalized image coordinates. The target point in the $D_1$ camera frame is then 

\begin{equation}
{}^{1}\mathbf{X}_{t}
=
Z_{t,1}
\begin{bmatrix}
x_{t,1}^{u}\\
y_{t,1}^{u}\\
1
\end{bmatrix}
\label{eq:target_backprojection}
\end{equation}

Using the fixed rigid transformation between the $D_1$ camera and the $D_1$-anchored virtual frame, this point is expressed in $\mathcal{F}_v$ as 

\begin{equation}
{}^{v}\mathbf{X}_{t}
=
\mathbf{R}_{v1}\,{}^{1}\mathbf{X}_{t}
+
\mathbf{t}_{v1}
\label{eq:target_virtual}
\end{equation}

where $\mathbf{R}_{v1}$ and $\mathbf{t}_{v1}$ are fixed by the mounting geometry of the $D_1$ camera relative to the virtual-frame origin. $D_1$ then transmits the resulting 3D target point ${}^{v}\mathbf{X}_{t}$ to $D_2$.

Upon receiving this point, $D_2$ transforms it into its own camera frame using the relative pose established above:

\begin{equation}
{}^{2}\mathbf{X}_{t}
=
\mathbf{R}_{2v}\,{}^{v}\mathbf{X}_{t}
+
\mathbf{t}_{2v}
=
\begin{bmatrix}
X_{t,2} & Y_{t,2} & Z_{t,2}
\end{bmatrix}^{T}
\label{eq:target_d2}
\end{equation}

The corresponding normalized image coordinates are

\begin{equation}
x_{t,2}=\frac{X_{t,2}}{Z_{t,2}},
\qquad
y_{t,2}=\frac{Y_{t,2}}{Z_{t,2}}
\label{eq:d2_normalized}
\end{equation}

To retain the calibrated camera model, radial and tangential distortion are then applied. With $r^2=x_{t,2}^{2}+y_{t,2}^{2}$,

\begin{equation}
\resizebox{\linewidth}{!}{$
\begin{bmatrix} x^d_{t,2} \\ y^d_{t,2} \end{bmatrix} = 
(1 + k_1 r^2 + k_2 r^4 + k_3 r^6) 
\begin{bmatrix} x^d_{t,2} \\ y^d_{t,2} \end{bmatrix} + 
\begin{bmatrix} 
2p_1 x_{t,2} y_{t,2} + p_2(r^2 + 2x^2_{t,2}) \\ 
p_1(r^2 + 2y^2_{t,2}) + 2p_2 x_{t,2} y_{t,2} 
\end{bmatrix}
$}
\end{equation}

Finally, the predicted target location in the $D_2$ image is 

\begin{equation}
\begin{bmatrix}
\hat{u}_{t,2}\\
\hat{v}_{t,2}\\
1
\end{bmatrix}
=
\mathbf{K}_2
\begin{bmatrix}
x_{t,2}^{d}\\
y_{t,2}^{d}\\
1
\end{bmatrix}
\label{eq:d2_projection}
\end{equation}

The projected pixel $(\hat{u}_{t,2},\hat{v}_{t,2})$ is used as a spatial prior indicating where the target is expected to appear in the receiver image. $D_2$ then performs its own target detection at this predicted location and generates a candidate bounding box $B_2^{c}$. Importantly, PATH does not geometrically generate or transmit a projected bounding box; the bounding box returned by $D_2$ is produced independently by its local detector. This candidate observation is transmitted back to $D_1$ and used by the Mutual Agreement Handshake described in Section~\ref{sec:handshake}.

\subsection{Mutual Agreement Handshake}
\label{sec:handshake}

The Mutual Agreement Handshake provides a lightweight cross-view verification mechanism before tracking responsibility is transferred from $D_1$ to $D_2$. Rather than relying on feature-level appearance matching between the two views, PATH verifies the geometric agreement between the target currently tracked by $D_1$ and the candidate independently detected by $D_2$.

As described in Section~\ref{virtual_3D}, $D_1$ first transmits the target's 3D location in the handoff-local frame $\mathcal{F}_v$. $D_2$ projects this point into its own image and invokes its local detector around the predicted pixel location. If a valid target candidate is detected at frame $k$, the detector produces a bounding box

\begin{equation}
B_{2,k}^{c}
=
(x_{2,k},y_{2,k},w_{2,k},h_{2,k})
\label{eq:d2_candidate}
\end{equation}

which is returned to $D_1$. No projected bounding box is generated by the geometric stage; the returned box is produced independently by the receiver-side detector.

To compare the two observations in a common image frame, $D_1$ expresses the receiver-side candidate in its own image coordinates. PATH transforms the reference corner $(x_{2,k},y_{2,k})$ using the inverse of the handoff-local geometric mapping described in Section~\ref{virtual_3D}, yielding $(\tilde{x}_{1,k},\tilde{y}_{1,k})$. Rather than reprojecting all four bounding-box corners independently, the box dimensions are adjusted according to the relative image scale between the two observations.

Under the calibrated pinhole model, the apparent image dimensions of the same physical target scale proportionally with focal length and inversely with viewing depth~\cite{10.1145/3550485}. Assuming a locally constant target depth across the bounding-box extent, the transformed width and height are therefore approximated as

\begin{equation}
\tilde{w}_{1,k}
=
w_{2,k}
\frac{f_{x,1}}{f_{x,2}}
\frac{z_{2,k}}{z_{1,k}},
\qquad
\tilde{h}_{1,k}
=
h_{2,k}
\frac{f_{y,1}}{f_{y,2}}
\frac{z_{2,k}}{z_{1,k}}
\label{eq:bbox_rescaling}
\end{equation}

where $z_{1,k}$ and $z_{2,k}$ denote the target viewing depths in the $D_1$ and $D_2$ camera frames, respectively. These quantities are available from the handoff-local geometry established in the preceding stage, and therefore no additional depth value needs to be transmitted with $B_{2,k}^{c}$. When the two cameras use identical intrinsic parameters, the focal-length ratios in (\ref{eq:bbox_rescaling}) reduce to unity.

The receiver observation expressed in the $D_1$ image is therefore

\begin{equation}
\widetilde{B}_{2\rightarrow1,k}^{c}
=
(\tilde{x}_{1,k},
 \tilde{y}_{1,k},
 \tilde{w}_{1,k},
 \tilde{h}_{1,k})
\label{eq:transformed_candidate}
\end{equation}

Let $B_{1,k}^{T}$ denote the active target bounding box maintained by $D_1$. PATH defines the cross-view verification score as

\begin{equation}
S_k
=
IoU
\left(
B_{1,k}^{T},
\widetilde{B}_{2\rightarrow1,k}^{c}
\right)
=
\frac{
\left|
B_{1,k}^{T}
\cap
\widetilde{B}_{2\rightarrow1,k}^{c}
\right|
}{
\left|
B_{1,k}^{T}
\cup
\widetilde{B}_{2\rightarrow1,k}^{c}
\right|
}
\label{eq:iou_score}
\end{equation}

Using the complete bounding-box overlap incorporates both positional and spatial-extent agreement between the two observations, rather than relying only on a centroid displacement. However, the same image-plane localization error can produce different IoU values at different apparent target scales. PATH therefore adjusts the acceptance threshold according to the observed target scale.

To guard against transient detections and sensing errors, an instantaneous verification indicator is defined at each frame $k$ as

\begin{equation}
q_k =
\begin{cases}
1, & \text{if } S_k \geq \tau_k,\\
0, & \text{otherwise}.
\end{cases}
\label{eq:instant_verification}
\end{equation}

If $D_2$ does not return a valid candidate detection at frame $k$, $q_k$ is also set to zero. The handshake is completed only when the required agreement is maintained continuously over the prescribed temporal window.

PATH uses an adaptive IoU acceptance threshold to account for changes in apparent target scale and bounding-box size. Let $A_{b,k}$ denote the representative box area in the $D_1$ image, computed as

\begin{equation}
A_{b,k}
=
\frac{
|B_{1,k}^{T}|
+
|\widetilde{B}_{2\rightarrow1,k}^{c}|
}{2}
\label{eq:representative_area}
\end{equation}

The expected target pixel area is modelled according to the inverse
square relationship with viewing distance under the pinhole camera model~\cite{9409130}:

\begin{equation}
A_t(z_k)
=
A_t(z_0)
\left(
\frac{z_0}{z_k}
\right)^2
\label{eq:expected_area}
\end{equation}

where $A_t(z_0)$ is the reference target area at viewing distance $z_0$, and $z_k$ is the current target viewing distance used for scale normalization. A bounded scale factor is then defined by normalizing the expected apparent target area with respect to the reference scale:

\begin{equation}
s_k
=
\mathrm{clip}
\left(
\left(\frac{z_0}{z_k}\right)^2,
0,
1
\right)
\label{eq:scale_factor}
\end{equation}

The adaptive acceptance threshold is

\begin{equation}
\tau_k
=
\tau_{\min}
+
(\tau_{\max}-\tau_{\min})s_k
\label{eq:adaptive_threshold}
\end{equation}

The threshold is therefore highest when the target is observed at or above the reference scale and is progressively relaxed as the apparent target size decreases with increasing viewing distance. A handoff is accepted only when $S_k \geq \tau_k$ is maintained for $N$ consecutive frames. The temporal verification decision is expressed as

\begin{equation}
V_k =
\begin{cases}
1, & \text{if } q_i = 1
\text{ for all }
i \in \{k-N+1,\ldots,k\},\\
0, & \text{otherwise}.
\end{cases}
\label{eq:temporal_verification}
\end{equation}

Any frame for which the IoU falls below the adaptive threshold, or for which no valid receiver-side candidate is detected, resets the consecutive-agreement condition. Once the temporal agreement criterion is satisfied, tracking responsibility is transferred from $D_1$ to $D_2$.

\subsection{Sensor Sensitivity}
\label{sec:sensitivity}

The preceding receiver-view projection depends primarily on the metric target depth measured by $D_1$ and the relative pose estimated from the UAV-mounted fiducial marker. To examine how errors in these inputs affect the projected target location in the $D_2$ image, we propagate literature-reported depth-measurement errors for the Intel RealSense D415 and fiducial-based translation and rotation errors through a nominal receiver-view projection. This analysis characterises geometric sensitivity; it is not presented as an empirical accuracy bound or a calibrated uncertainty distribution for the deployed PATH system.

For this analysis, a nominal camera matrix is derived from the manufacturer-reported D415 RGB resolution and field of view. Using $f_x=\frac{W}{2\tan(\theta_x/2)},~ f_y=\frac{H}{2\tan(\theta_y/2)}$ with $W=1920$, $H=1080$, $\theta_x=69.4^{\circ}$, and $\theta_y=42.5^{\circ}$~\cite{intelrealsense2019} gives

\begin{equation}
\mathbf{K}_{\mathrm{nom}}=
\begin{bmatrix}
1386.42 & 0 & 960\\
0 & 1388.61 & 540\\
0 & 0 & 1
\end{bmatrix}
\label{eq:nominal_intrinsics}
\end{equation}

The calculation assumes rectified image coordinates and therefore uses a zero-distortion nominal pinhole model. This simplification is used only for the sensitivity analysis; the deployed PATH projection retains the calibrated lens-distortion model described in Section~\ref{virtual_3D}. Importantly, $\mathbf{K}_{\mathrm{nom}}$ defines only the nominal sensitivity-analysis geometry and is not presented as the device-specific calibration of the camera deployed in the experiments. A separate intrinsic-calibration perturbation is not introduced because an intrinsic-parameter covariance or reprojection-error distribution is unavailable for the deployed unit.

\begin{table}[!h]
\centering
\caption{Literature-based reference errors used in the sensitivity analysis.}
\label{tab:reference_sensor_errors}
\footnotesize
\setlength{\tabcolsep}{3pt}
\renewcommand{\arraystretch}{1.05}
\begin{tabular}{@{}p{0.48\columnwidth}p{0.43\columnwidth}@{}}
\hline
\textbf{Input} & \textbf{Reference error}\\
\hline

Target depth (Intel RealSense D415)
\cite{lourencco2021intel, servi2024comparative}
&
5~mm RMSE at 1~m\\

AprilTag translation \cite{pfleging2015dynamic}
&
4.3~mm mean (3.2~mm SD)\\

AprilTag rotation \cite{pfleging2015dynamic}
&
$1.83^{\circ}$ mean ($1.77^{\circ}$ SD)\\
\hline
\end{tabular}
\end{table}

The cited marker experiment used a 58-mm AprilTag over a distance of 0.8--1.2~m, whereas PATH uses a 40-mm marker over an operating distance of 0.6--1.5~m. The cited values are therefore treated only as reference magnitudes for sensitivity analysis and not as measured error bounds for the PATH implementation.

The target-depth perturbation, based on the reported D415 measurement error, is evaluated with both positive and negative signs. For the fiducial-pose perturbations, translation directions and rotation axes are evaluated uniformly while retaining the reference magnitudes in Table~\ref{tab:reference_sensor_errors}. This directional evaluation is used to avoid assuming an unsupported per-axis sensor-error distribution. Accordingly, confidence intervals and percentile bounds are not inferred from the cited mean errors.

The direction-averaged RMS projection sensitivity is defined as

\begin{equation}
e_{\mathrm{RMS}}=
\sqrt{
    \frac{1}{M}
    \sum_{j=1}^{M}
    \left\|
        \hat{\mathbf{p}}_{2,j}
        -
        \mathbf{p}_{2}^{\mathrm{nom}}
    \right\|_2^2
}
\label{eq:projection_rms}
\end{equation}

where $\mathbf{p}_{2}^{\mathrm{nom}}$ is the nominal projected target location in the $D_2$ image and $\hat{\mathbf{p}}_{2,j}$ is the corresponding location obtained under perturbation direction $j$. The evaluation uses $M=5\times10^5$ directions. This sampling approximates an average over possible perturbation directions and should not be interpreted as a probabilistic sensor-error model. The combined sensitivity is reported as the root-sum-square of the independently evaluated target-depth and relative-pose sensitivities.

The nominal geometry places the sender-side target at the principal point at a depth of 1.0~m and uses

\begin{equation}
\mathbf{t}_{2v}
=
[-1.0,\;0.2,\;1.0]^\top\ \mathrm{m}
\end{equation}

with a relative yaw of $25^{\circ}$, where $\mathbf{t}_{2v}$ denotes the relative translation from the $D_1$-anchored handoff-local frame $\mathcal{F}_v$ to the $D_2$ camera frame. The resulting nominal receiver-view projection is $(540.08,\,685.69)$ pixels.

\begin{figure}[!t]
    \centering
    \includegraphics[width=0.8\linewidth]{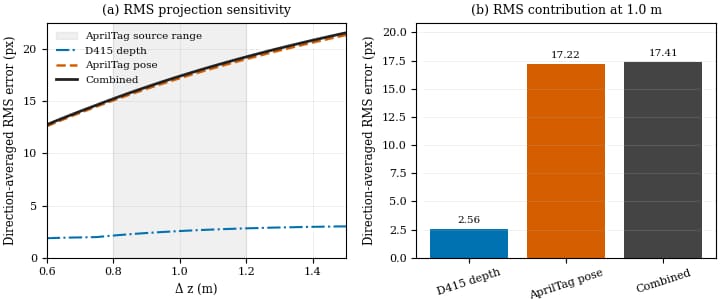}
    \caption{Receiver-view projection sensitivity to target-depth and
    fiducial-pose perturbations under the nominal handoff geometry.}
    \label{fig:sensitivity}
\end{figure}

At a target depth of 1.0~m, the target-depth and fiducial-marker relative-pose terms produce RMS receiver-view projection sensitivities of 2.56 and 17.22 pixels, respectively. Their combined sensitivity is 17.41 pixels, corresponding to approximately 0.79\% of the diagonal of the $1920\times1080$ receiver image. Across the evaluated 0.6--1.5~m target-depth range, the combined sensitivity increases from 12.75 to 21.57 pixels.

The fiducial-marker relative-pose term is therefore the dominant contributor under the evaluated nominal geometry. These results support using the projected target point as a spatial search prior rather than requiring exact point-to-point correspondence between the two UAV views. The subsequent region-based Adaptive-IoU verification provides tolerance to moderate projection and detection variation, while the repeated temporal-agreement condition prevents a single perturbed observation from immediately triggering a handoff.

%% file: Sections/Setup.tex
\section{Experimental Setup}
\label{subsec:system_setup}

To evaluate PATH across different UAV hardware platforms, we used two pairs of quadrotors: DJI Tello drones and custom 425~mm quadrotors equipped with Pixhawk Cube Orange flight controllers and NVIDIA Jetson Nano modules. The cameras on both platforms were calibrated individually using the same checkerboard-based procedure to obtain their intrinsic and lens-distortion parameters. The custom UAVs were equipped with Intel RealSense D415 RGB-D cameras mounted at approximately $45^{\circ}$, while the Tello platform used its onboard cameras with separately calibrated parameters. The sender UAV $D_1$ additionally carried a $40\times40$~mm fiducial marker for relative-pose estimation. The tracked targets were not instrumented.

The quantitative geometric and sensor-sensitivity evaluations were performed using the D415-equipped custom UAV platform; the Tello pair was used to evaluate PATH on a different UAV hardware platform. The D415-based sensitivity analysis in Section~\ref{sec:sensitivity} is therefore intended to characterise the sensing errors of the metric projection configuration rather than every camera platform.

Inter-UAV communication was performed over Wi-Fi using the compact geometric observations described in Section~\ref{virtual_3D} and Section~\ref{sec:handshake}. A handoff was accepted when the required cross-view agreement was maintained for $N=350$ consecutive frames at 50~Hz, corresponding to a 7~s verification window.

We evaluated PATH with three representative target types: (i) a 1/16-scale robot car, (ii) micro-UAVs, and (iii) a walking human. Experiments were conducted in an indoor flight facility with controlled lighting and no wind and in an outdoor field with natural illumination and background clutter. The indoor experiments additionally used an ultrasonic positioning system to evaluate the estimated UAV and target positions. The evaluation examines receiver-view projection accuracy, cross-view verification reliability, performance against appearance-based matching baselines, and runtime and communication overhead.

A total of six volunteers participated in the data collection. This study was approved by the Human Research Ethics Committee  (HREC) of Macquarie University (approval no. 520262029268789), and written informed consent was obtained from all subjects prior their participation.

%% file: Sections/Results.tex
\section{Results}
\subsection{Receiver-View Projection Consistency}
\label{sec:projection_consistency}

\begin{figure}[!h]
    \centering
    \begin{subfigure}[b]{0.485\linewidth}
        \centering
        \includegraphics[width=\linewidth]{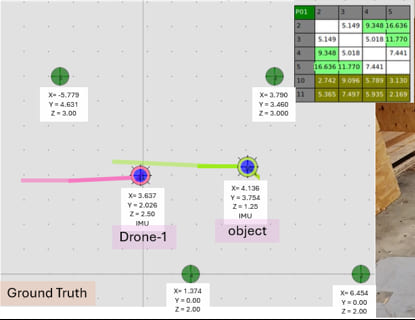}
        \caption{Ground-truth map from ultrasonic beacons}
        \label{fig:gt_map}
    \end{subfigure}
    \hfill
    \begin{subfigure}[b]{0.485\linewidth}
        \centering
        \includegraphics[width=\linewidth]{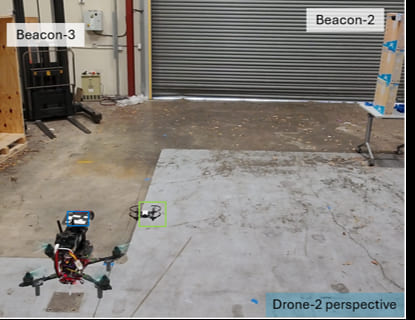} 
        \caption{D2’s perspective detecting D1’s marker and target}
        \label{fig:drone2_cam}
    \end{subfigure}
    \caption{Indoor positioning experiment}
    \label{ind-pos}
\end{figure}

We first evaluated the handoff-local representation under dynamic motion using an ultrasonic indoor positioning system as an independent trajectory reference. The ultrasonic measurements were time-synchronized with the onboard UAV logs and used to compare the motion estimated by PATH with the measured UAV and target trajectories. As shown in Fig.~\ref{ind-pos}, the receiver UAV $D_2$ continuously observes the fiducial marker on $D_1$ and uses the transferred target location to identify the target in its own view while the UAVs and target are in motion.

\begin{figure}[!ht] 
    \centering
    \begin{subfigure}[b]{0.44\linewidth}
        \centering
        \includegraphics[width=\linewidth]{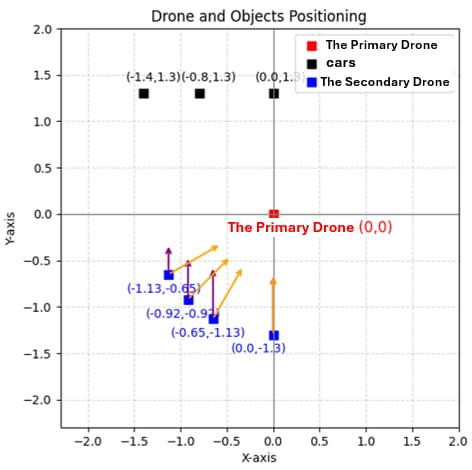} 
        \caption{Coordinate frame}
        \label{fig:grouped_a}
    \end{subfigure}
    \hfill
    \begin{subfigure}[b]{0.54\linewidth}
        \centering
        \includegraphics[width=\linewidth]{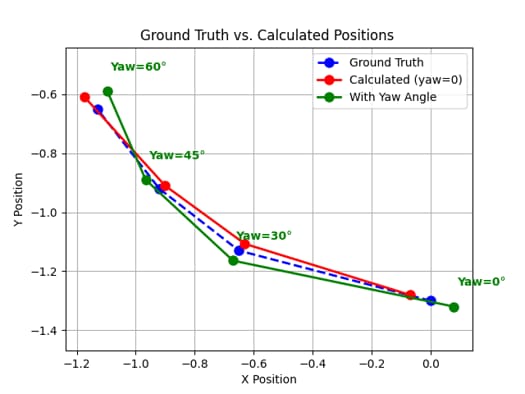} 
        \caption{Ground-truth}
        \label{fig:grouped_b}
    \end{subfigure}

    \caption{Relative positioning analysis}
    \label{drone2_Pos}
\end{figure}

We separately quantified the geometric accuracy under controlled static conditions. Seven relative $D_2$ configurations, including different positions and orientations around $D_1$, were evaluated, with three target locations incorporated across these cases, as illustrated in Fig.~\ref{fig:grouped_a}. Each $D_2$ positioning case was measured 60 times, resulting in a total of 420 measurements. The physical relative geometry was measured manually and used as the reference for comparison with the PATH estimates.

Across the static evaluation, the mean positional error of $D_2$ relative to $D_1$ was $0.047\,\mathrm{m}$, while the mean target-position error was $0.030\,\mathrm{m}$. Figure~\ref{fig:grouped_b} further compares the estimated $D_2$ position under different relative orientations. When $D_2$ faced in the same direction as $D_1$, the mean positional error was $0.038\,\mathrm{m}$. When $D_2$ faced toward $D_1$, introducing a relative yaw angle, the mean error increased to $0.054\,\mathrm{m}$. This increase is consistent with the higher sensitivity to fiducial-based relative-pose perturbations identified in Section~\ref{sec:sensitivity}.

Together, the two experiments examine complementary aspects of the handoff local representation. The dynamic experiment demonstrates that the transferred spatial information remains consistent during motion, while the static evaluation quantifies the positional discrepancy under controlled relative UAV and target configuration.

\subsection{Receiver-Side Target Acquisition under Visual Ambiguity}
\label{sec:target_aq}
Having validated the handoff local geometric representation achieve cm-scale accuracy, we next evaluate how reliably the receiver UAV $D_2$ can acquire the intended target before the Mutual Agreement Handshake. This stage is critical because an incorrect receiver side candidate may correspond to a nearby distractor rather than the target currently tracked by $D_1$.

We compare PATH with two appearance based feature matching baselines: ORB~\cite{rublee2011orb}, representing classical hand crafted local features, and XFeat~\cite{potje2024cvpr}, representing modern lightweight deep learning (DL) based local features. These were chosen to ensure a comprehensive comparison across both traditional pre-DL and modern DL feature matching techniques. Conversely, XFeat represents the state of art in lightweight, learning-based feature extraction, providing enhanced robustness to illumination and viewpoint changes while maintaining real-time capabilities as well as openly available source code. ORB and XFeat attempt to identify the corresponding target in the receiver view from visual appearance, whereas PATH uses the 3D perspective matching and target location transferred by $D_1$ and projected into the $D_2$ image as a spatial prior for local target detection.

To examine conditions in which appearance alone becomes ambiguous, the outdoor experiment included six individuals wearing similar clothing and caps. This created multiple visually similar candidates in the receiver view, increasing the likelihood of incorrect association or failed correspondence with the intended target. 

\begin{figure}[!htbp]
    \centering
    \includegraphics[width=0.7\linewidth]{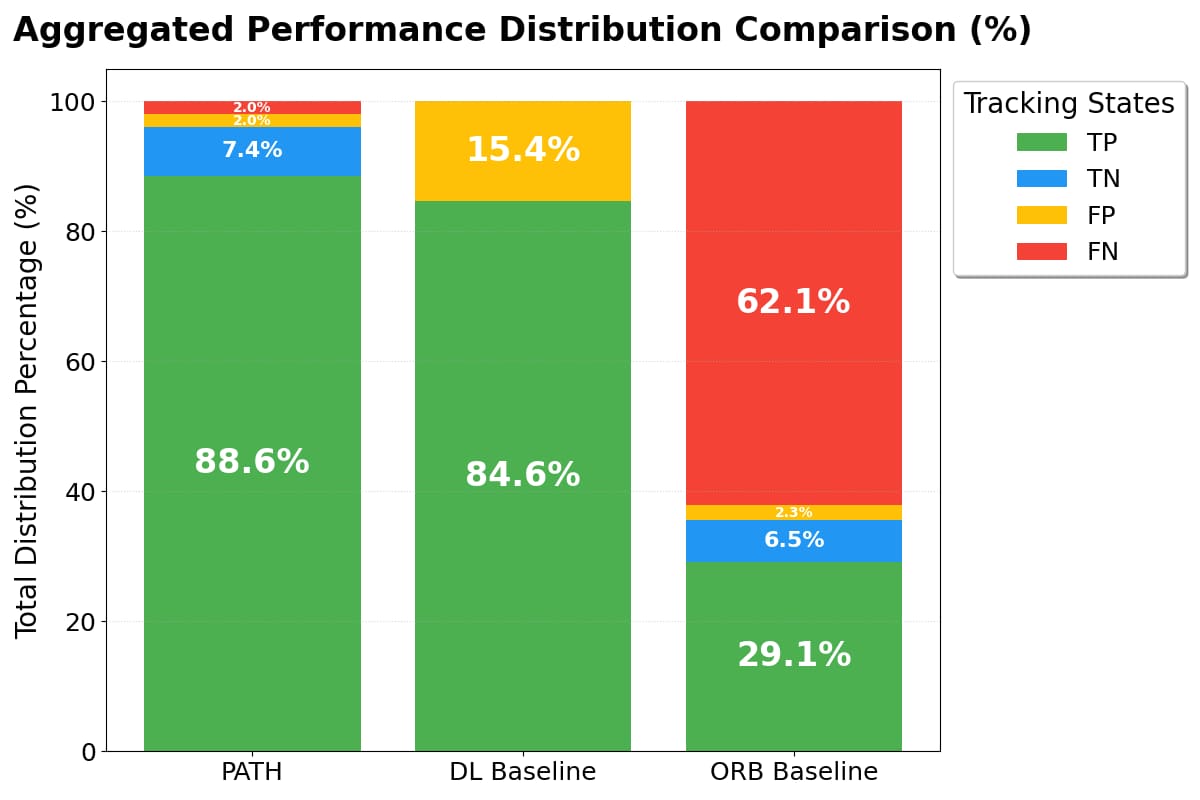}
    \caption{Frame-level receiver-side target-acquisition outcomes in
    the visually ambiguous outdoor experiments.}
    \label{fig:aggregated_perf}
    \vspace{-5pt}
\end{figure}

\begin{figure*}[htbp]
    \centering
    \hfill
    \begin{subfigure}[c]{0.3\textwidth}
        \centering
        \includegraphics[width=\textwidth]{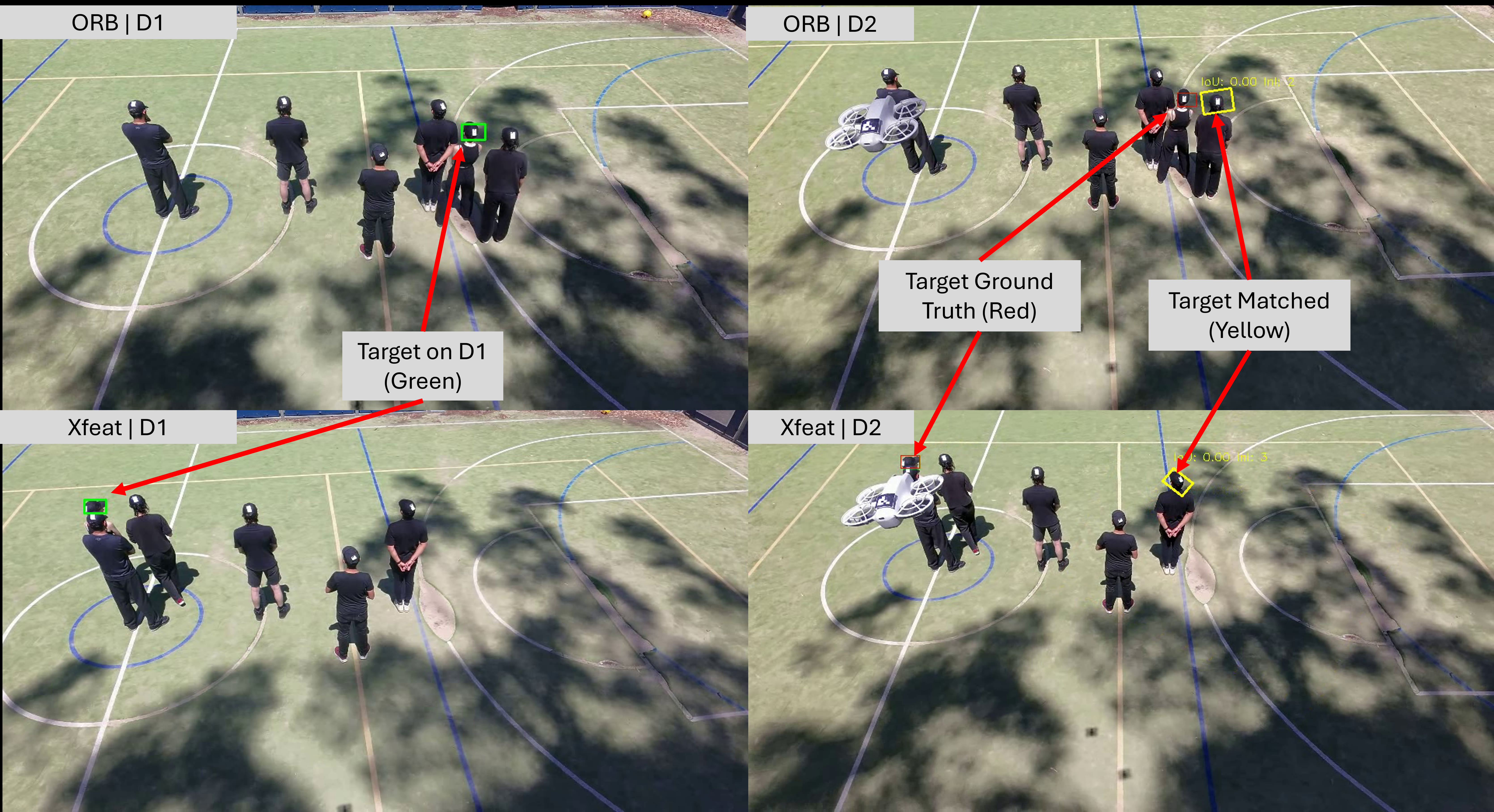}
        \caption{False Positive: i) ORB ii) XFeat}
        \label{fig:error_a}
    \end{subfigure}
    \hfill
    \begin{subfigure}[c]{0.32\textwidth}
        \centering
        \includegraphics[width=\textwidth]{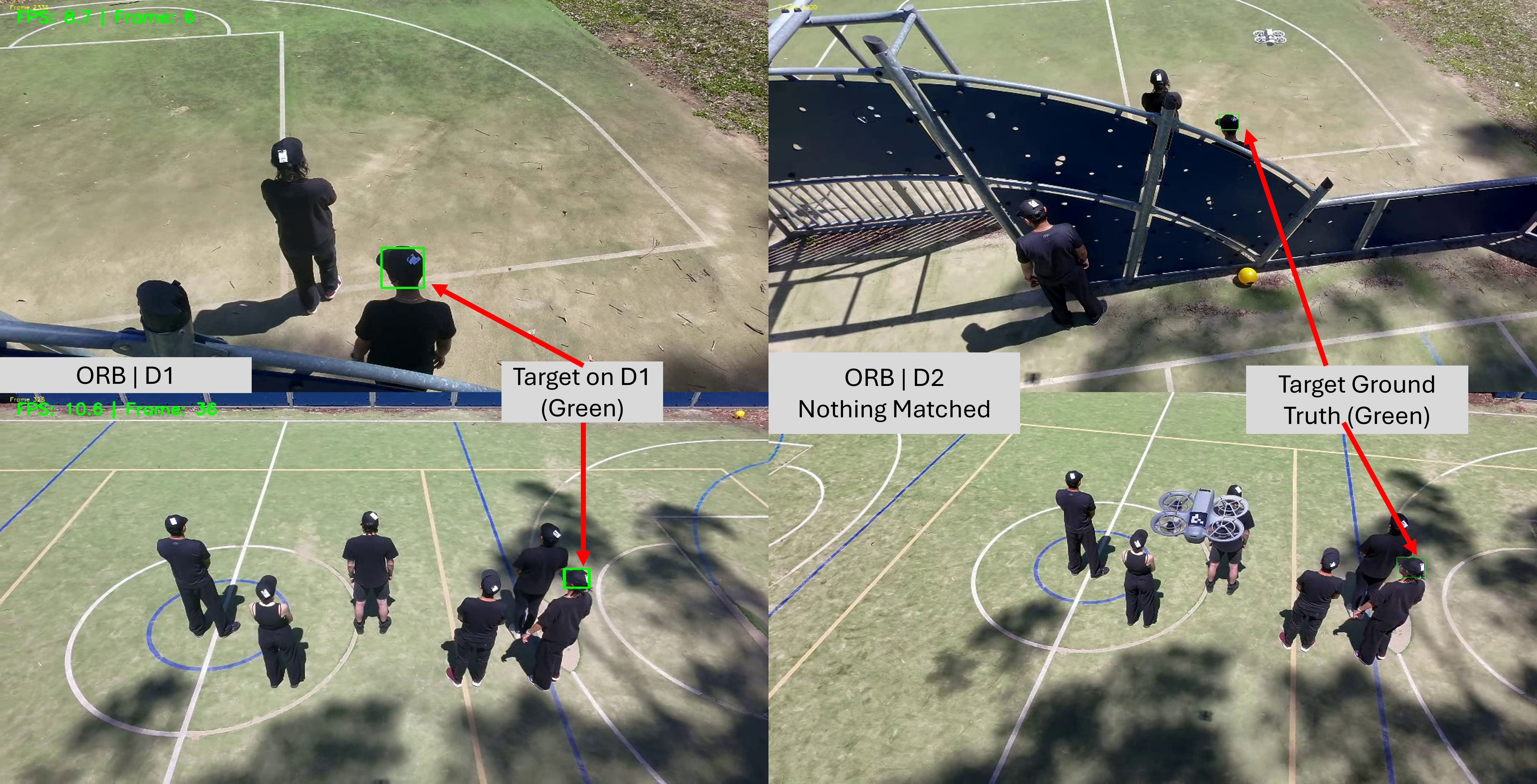}
        \caption{ORB False Negative}
        \label{fig:error_b}
    \end{subfigure}
    \hfill
    \begin{subfigure}[c]{0.28\textwidth}
        \centering
        \includegraphics[width=\textwidth]{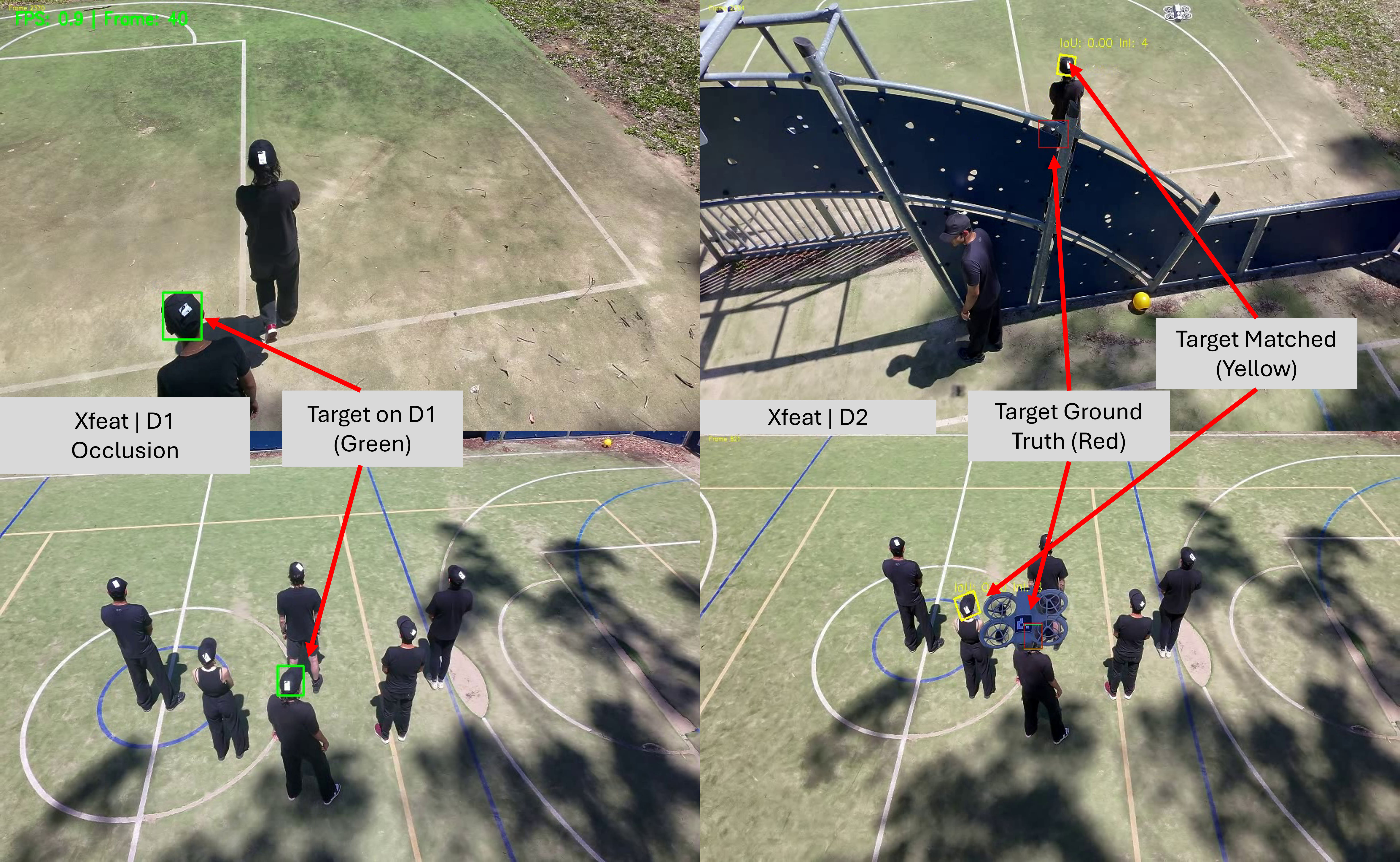}
        \caption{XFeat False Positive}
        \label{fig:error_c}
    \end{subfigure}
    \caption{Comparison of feature matching failures cases.}
    \label{fig:Error cases}
\end{figure*}

Figure~\ref{fig:aggregated_perf} summarizes the frame-level receiver-side target-acquisition outcomes across the evaluated outdoor sequences, while Fig.~\ref{fig:Error cases} presents representative failure cases. Each frame is classified according to whether $D_2$ correctly acquires the intended target specified by $D_1$, selects an incorrect candidate, or fails to acquire a valid target. 

The evaluation comprised 350 frame-level observations, including 317 positive cases in which the sender-specified target was present in the receiver's view and 33 negative cases in which no valid correspondence with the sender-specified target should be accepted. A true positive (TP) was recorded when the intended target was present, and $D_2$ correctly acquired it. A false positive (FP) occurred when $D_2$ selected an incorrect candidate, such as a visually similar distractor. A false negative (FN) was recorded when the intended target was present but was not acquired. A true negative (TN) was recorded when no valid correspondence existed, and the method correctly rejected the available receiver-side candidates.

ORB exhibited a high false-negative rate of 62.1\%, indicating frequent failure to establish sufficient correspondence with the intended target under the evaluated viewpoint and appearance variations. XFeat was more robust to correspondence loss but produced a false-positive rate of 15.4\%, with several failures corresponding to visually similar distractors being selected instead of the intended target.

PATH produced 310 true-positive, 26 true-negative, 7 false-positive, and 7 false-negative observations, corresponding to 88.6\%, 7.4\%, 2.0\%, and 2.0\% of the evaluated frames, respectively. This gives an overall frame-level target-acquisition accuracy of 96.0\%. The low false-positive and false-negative rates indicate that the spatial prior improves discrimination among visually similar receiver-side candidates under the evaluated conditions. 


Figure~\ref{fig:error_a} shows cases in which the appearance-based methods associate the observation from $D_1$ with an incorrect individual in the $D_2$ view, resulting in false positives. Figure~\ref{fig:error_b} shows an ORB false negative in which the intended target remains visible but sufficient correspondence is not established. Figure~\ref{fig:error_c} illustrates an XFeat failure under degraded target visibility, where correspondence shifts to an incorrect candidate.

These results do not imply that appearance-based matching is generally ineffective. When the target remains visually distinctive, both appearance-based methods can establish reliable correspondence. The advantage of PATH becomes more apparent when similar distractors, viewpoint variation, or degraded visibility make appearance-based association ambiguous. In these cases, the transferred geometric prior narrows the receiver-side search to the expected target location before the candidate is passed to the Mutual Agreement Handshake.

\subsection{Hardware Performance}
\label{sec:Hardware}

All handoff computation runs CPU-only on the Jetson Nano (quad-core A57, 4\,GB RAM, 5--10\,W envelope). On $D_1$, the handoff computation includes target depth retrieval and 3D back-projection, transformation of the target point into the handoff-local frame $\mathcal{F}_v$, and the bounding-box transformation and IoU calculation used for the Mutual Agreement Handshake. On $D_2$, the CPU performs fiducial-based relative pose estimation using PnP, transforms the received 3D target point into the $D_2$ camera frame, and reprojects it into the receiver image. The relative translation and orientation required for this transformation are therefore estimated locally by $D_2$ rather than transmitted by $D_1$. The geometric operations consist primarily of small matrix--vector operations and image-plane projections; in practice, they execute comfortably in real time at video rate (30--60\,Hz) without requiring an additional GPU-based feature-matching stage, leaving the GPU available for the onboard detector.

Inter-UAV messaging is likewise tiny: each cycle $D_1\!\to D_2$
transmits the reconstructed 3D target point ${}^{v}\mathbf{X}_t=(X_t,Y_t,Z_t)$ in the handoff-local frame, while $D_2\!\to D_1$ returns the detector-generated candidate bounding box $B_2^{c}=(x_2,y_2,w_2,h_2)$. Camera calibration parameters are stored locally, and the relative pose of $D_2$ with respect to $\mathcal{F}_v$ is computed onboard from the fiducial observation, so these quantities do not need to be transmitted at every cycle. With 32-bit fields, a packet is $\leq\!128$\,B, so even bidirectional exchange at 60\,Hz is $<\!16$\,kB/s---negligible on a commodity Wi-Fi link (tens--hundreds of Mb/s). The lightweight payload and $N$-frame IoU gating reduce communication overhead and support real-time operation without requiring transmission of raw images, depth frames, or visual descriptors. Across 420 trials, the system sustained real-time (30--60\,Hz) operation with no communication-induced stalls; the effective bidirectional payload averaged $\approx12$\,kB/s (typically 9--15\,kB/s), and message latency remained well below the frame interval, enabling consistently real-time handoffs.

\begin{figure}[!htpb]
    \centering
    \includegraphics[width=\linewidth]{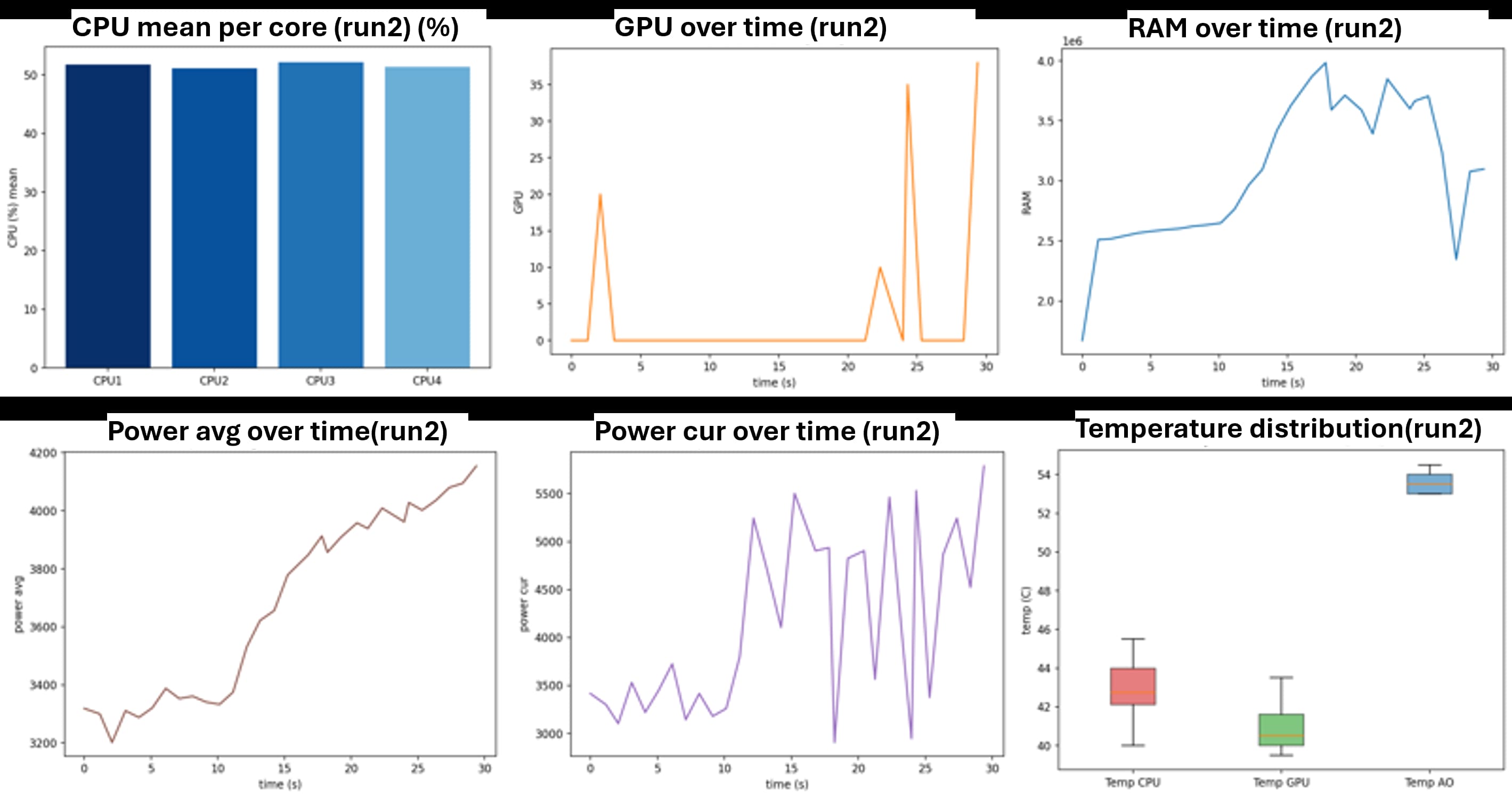}
    \caption{ORB pipeline hardware run} 
    \label{fig:orb-jtop}
    \vspace{-10pt}
\end{figure}
In contrast, feature-based handoff baselines impose a much heavier onboard load on the same platform, which directly translates to latency and timing jitter during handoff. Figure \ref{fig:orb-jtop} shows the ORB pipeline resource profile. CPU utilisation is high across all four cores throughout the run, while GPU use remains minimal. RAM usage increases during processing, and both instantaneous and average power rise relative to idle, with corresponding thermal uplift across CPU, GPU, and AO sensors. This indicates the baseline is primarily CPU-bound on Jetson Nano, leaving limited headroom for other real-time tasks (detection, tracking, control). Under these conditions, the system is prone to frame drops and delayed updates, which aligns with the poorer handoff stability observed for ORB compared with our method.
\begin{figure}[!htpb]
    \centering
    \includegraphics[width=\linewidth]{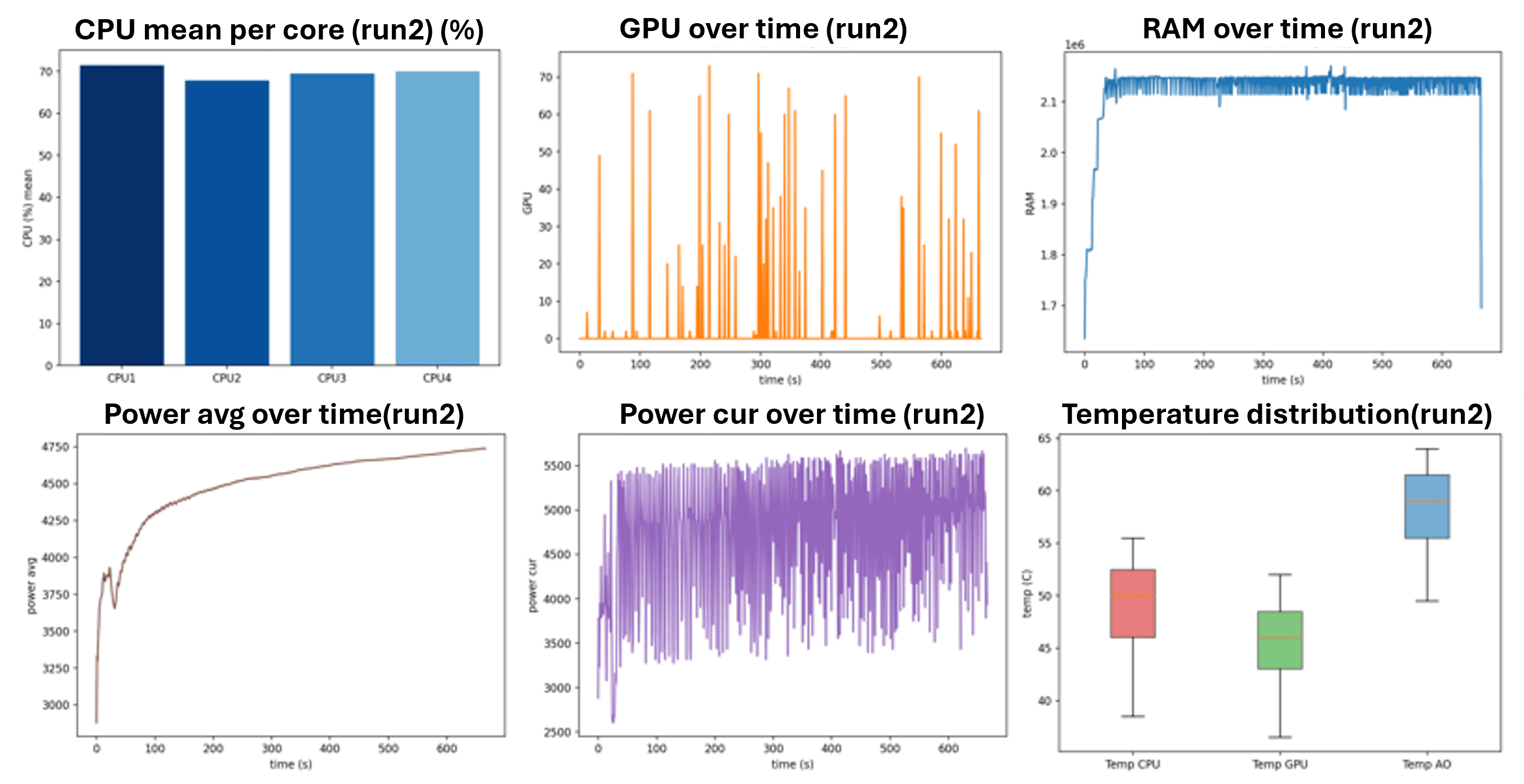}
    \caption{ XFeat pipeline hardware run} 
    \label{fig:xfeat-jtop}
\end{figure}

Figure \ref{fig:xfeat-jtop} shows the XFeat pipeline resource profile. Unlike ORB, XFeat triggers frequent GPU bursts during matching, alongside sustained CPU load. Power draw is higher and more variable, and the thermal distribution shifts upward, reflecting a heavier compute footprint and greater runtime variability. This bursty behaviour introduces jitter in the processing loop and reduces the ability to maintain steady real-time timing on embedded hardware, especially when the GPU is also needed for other perception modules. Taken together, the jtop profiles demonstrate why ORB and XFeat degrade handoff performance on Jetson Nano: both consume substantial compute and power budget, whereas our CPU-only geometric handoff keeps resource usage low and predictable, preserving real-time operation.

\subsection{Mutual Agreement Handshake}

\begin{figure}[!ht]
    \centering
    \begin{subfigure}[b]{0.24\linewidth}
        \centering
        \includegraphics[width=\linewidth]{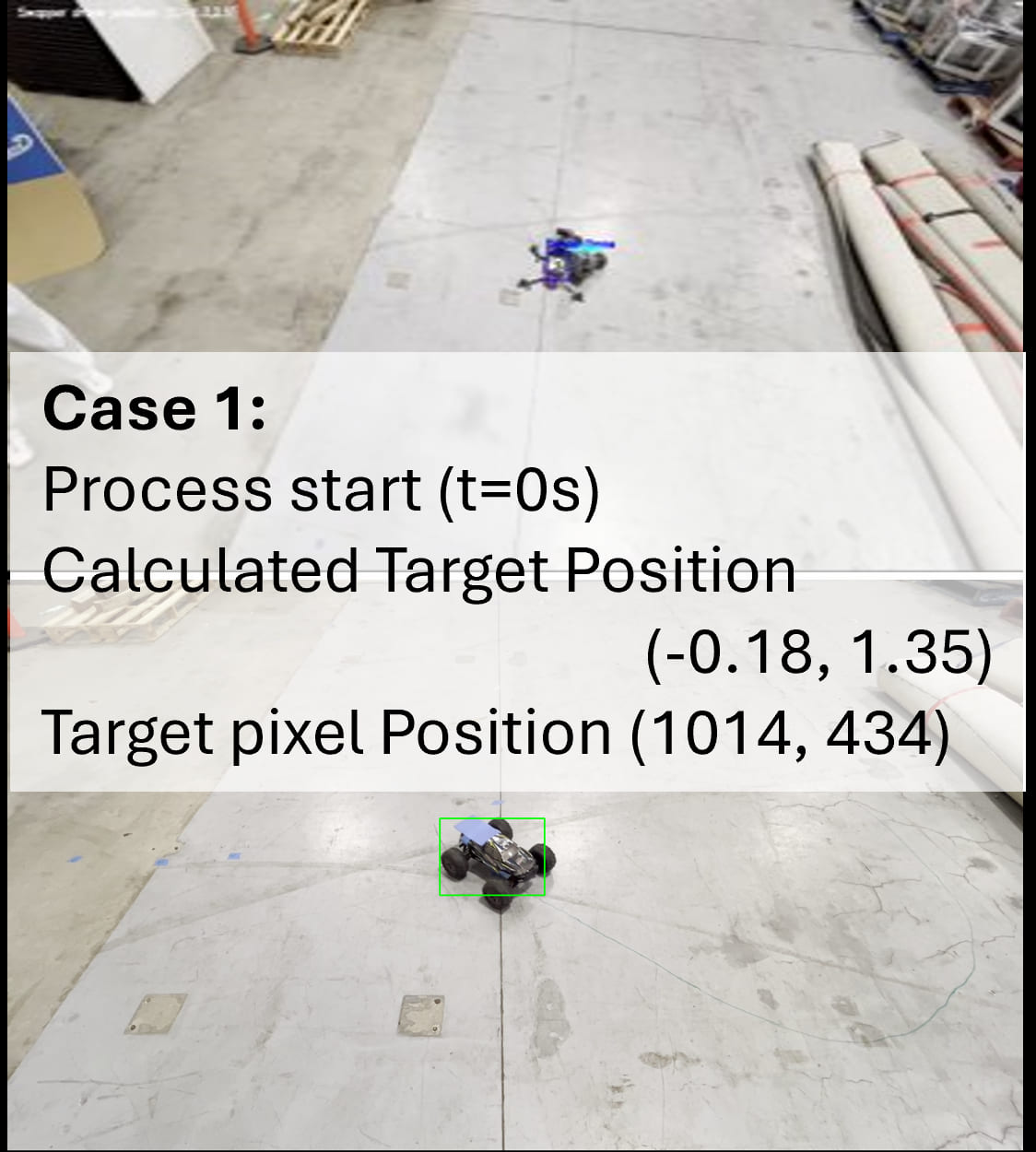}
        \caption{[t=0s]}
        \label{fig:process_1}
    \end{subfigure}
    \hfill
    \begin{subfigure}[b]{0.24\linewidth}
        \centering
        \includegraphics[width=\linewidth]{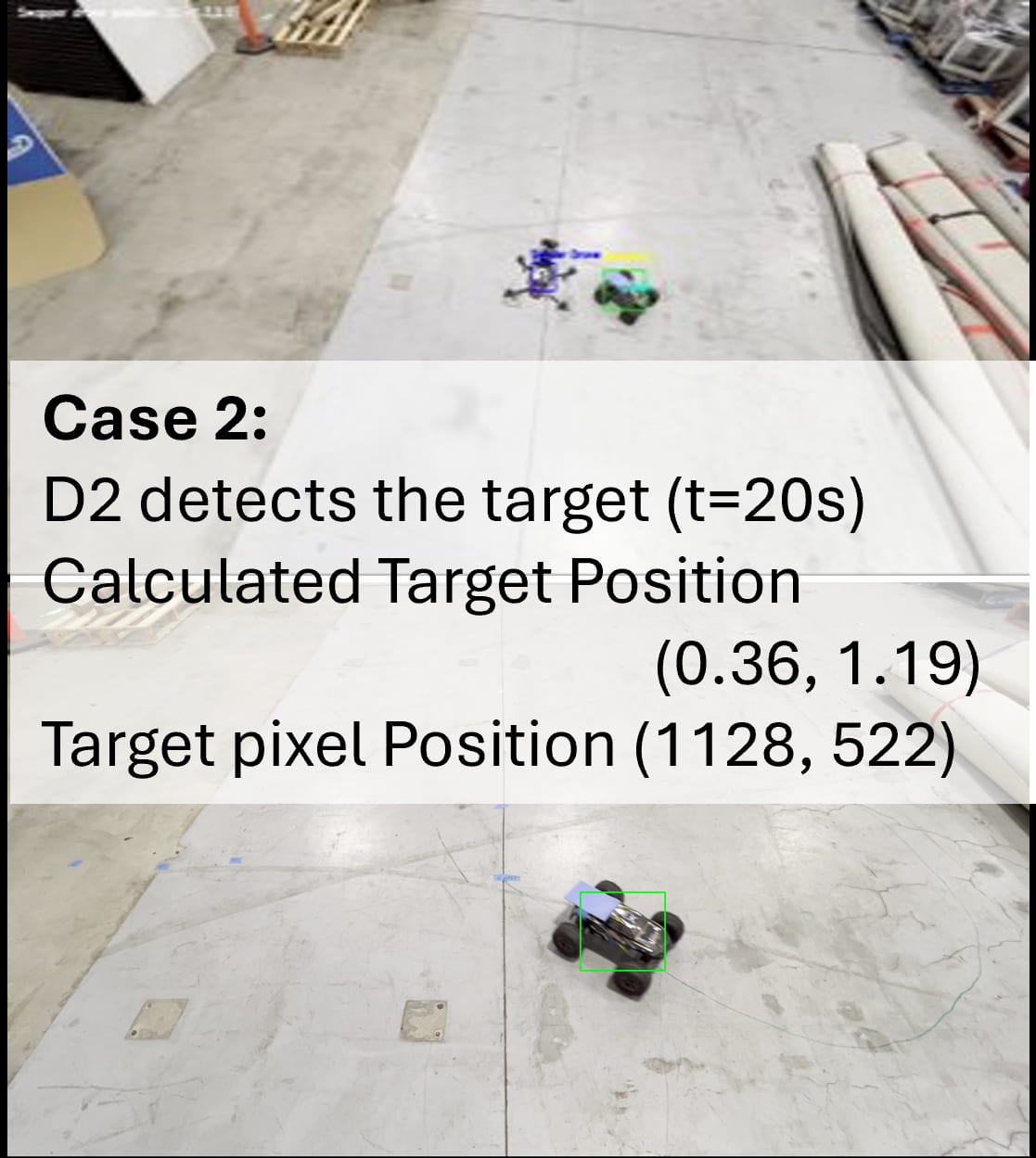}
        \caption{[t=20s]}
        \label{fig:process_2}
    \end{subfigure}
    \hfill
    \begin{subfigure}[b]{0.24\linewidth}
        \centering
        \includegraphics[width=\linewidth]{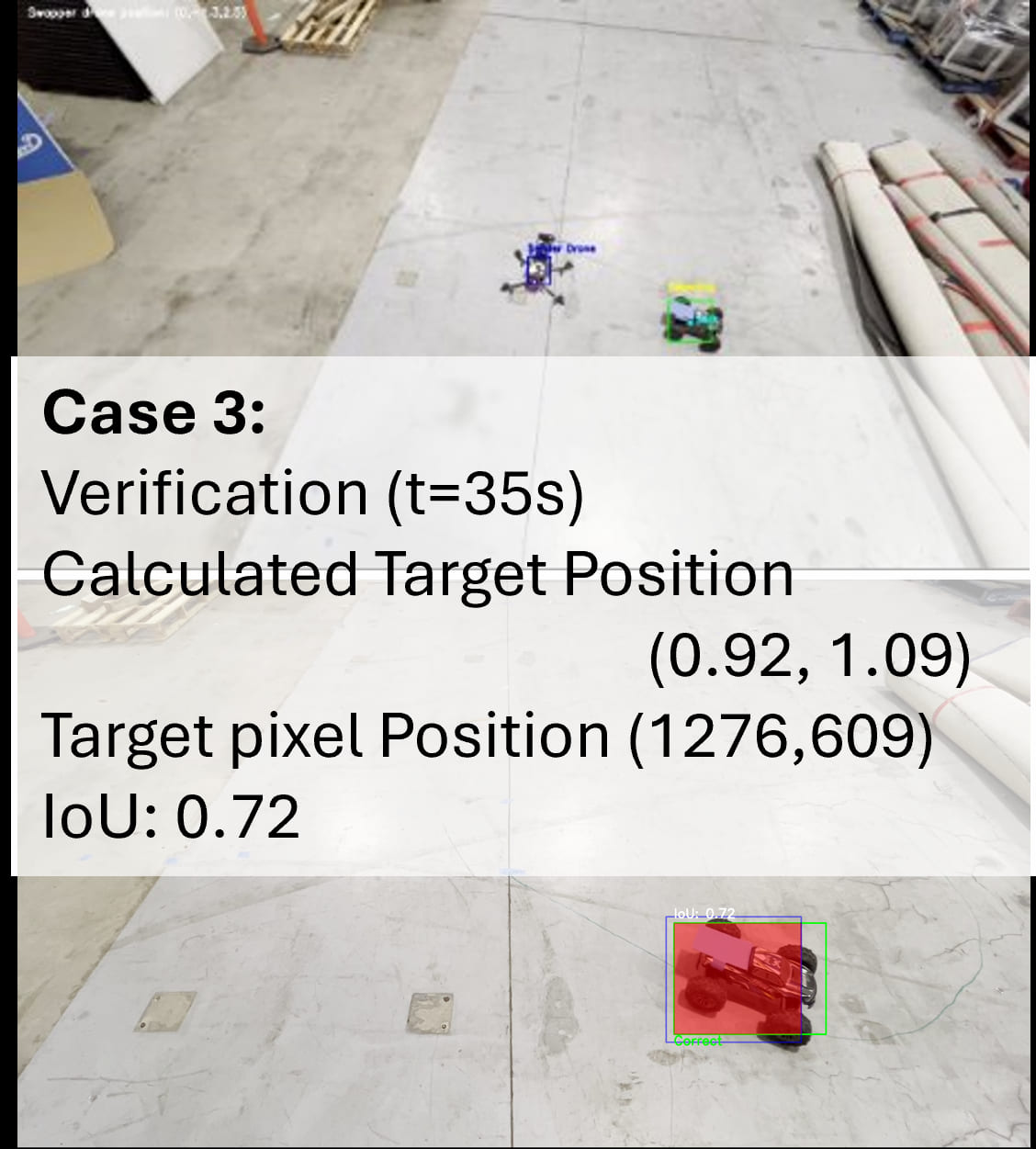}
        \caption{[t=35s]}
        \label{fig:process_3}
    \end{subfigure}
    \hfill
    \begin{subfigure}[b]{0.24\linewidth}
        \centering
        \includegraphics[width=\linewidth]{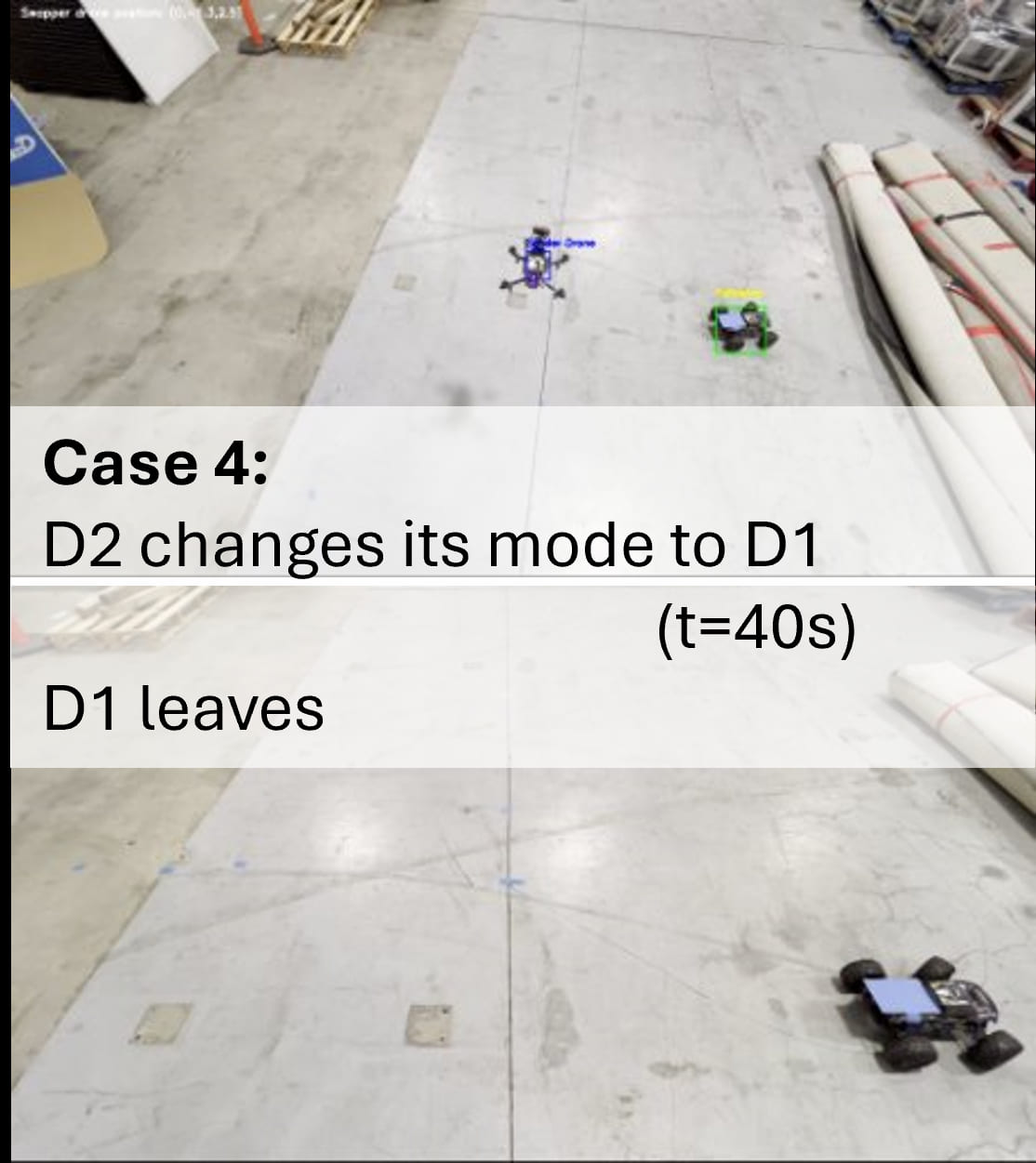}
        \caption{[t=40s]}
        \label{fig:process_4}
    \end{subfigure}
    
    \caption{Operational sequence: (a) Start, (b) Detect, (c) Verify, (d) Mode change.}
    \label{totalProcess}
\end{figure}

Finally, we evaluate the complete handoff sequence from receiver side target acquisition to the transfer of tracking responsibility. Fig. \ref{totalProcess} provides a detailed, step-by-step visualization of the tasks executed by the D1 (lower panel) and D2 (upper panel) at each stage. This visual breakdown clearly illustrates the coordinated interactions and synchronized efforts between the drones. At time $t=0$, once D2 detects D1, the operational sequence commences.

At the beginning of the sequence, $D_2$ observes the fiducial marker on $D_1$ and establishes the handoff-local relative geometry. After receiving the 3D target location from $D_1$, $D_2$ projects the target location into its own image and acquires a receiver-side candidate. The detected candidate bounding box is then returned to $D_1$, where it is transformed into the $D_1$ image frame and evaluated using the Mutual Agreement Handshake described in Section~\ref{sec:handshake}.

Once the cross-view agreement satisfies the acceptance criterion for the required number of consecutive frames, the handoff is confirmed and tracking responsibility is transferred from $D_1$ to $D_2$.

\begin{figure}[!htpb]
    \centering
    \includegraphics[width=0.7\linewidth]{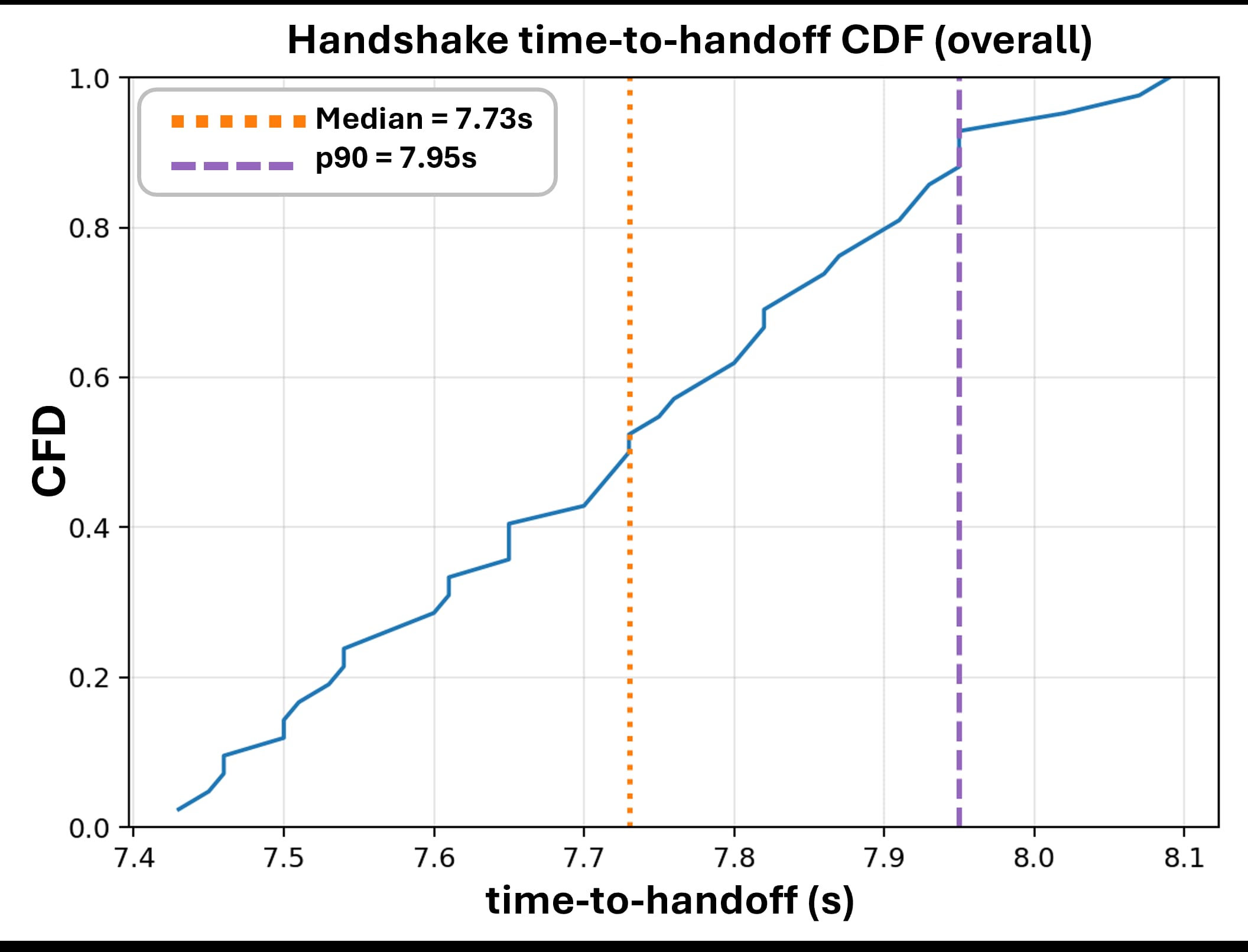}
    \vspace{-2pt}
    \caption{ Mutual Handshake performance}
    \label{fig:handshake}
    \vspace{-15pt}
\end{figure}

We quantify handshake completion time across all scenarios combined, using the 42 successful handoffs (out of 45 attempts; 3 timeouts). We measure time-to-handoff as the interval from the first frame in which IoU exceeds the acceptance threshold to the commit event, after maintaining a continuous 7 s hold at 50 Hz (350 frames). Fig.~\ref{fig:handshake} shows the distribution of $(t_{\mathrm{commit}}-t_{\mathrm{IoU_start}})$, with a median of 7.20 s and a 90th percentile of 7.95 s, indicating low timing jitter after agreement is reached. The tight distribution shows that once IoU exceeds the threshold, the commit time is consistent across trials, and differences in overall handoff timing mainly stem from how quickly D2 achieves initial agreement. The 3 timeouts correspond to attempts in which sustained IoU could not be maintained within the allotted window, thereby preventing an incorrect handoff.

%% file: Sections/discussion_conclusion.tex
\section{Discussion and Future Work}
\label{fut_work}


The current implementation has certain limitations. First, the relative geometry depends on reliable observation of the fiducial marker on $D_1$; temporary loss of this observation can interrupt receiver-view projection. 
Future work will therefore investigate markerless relative-pose estimation.  Second, objects in the environment may occlude the handoff process.  Hence, we plan to investigate methods for handling occlusion and other sources of temporary inter-UAV visibility loss through actively repositioning the receiver during handoff. 
Third, projection accuracy remains sensitive to relative-pose and depth errors, as shown by the sensitivity analysis. Also, the present evaluation considers two-UAV handoff under the tested indoor and outdoor conditions and does not address larger multi-UAV coordination.  We plan to explore extension of the handoff protocol to larger UAV teams and multi-UAV visibility planning, which may help reduce the projection error.
Further, PATH's design requires communication among drones to verify the handoff, making it vulnerable to jamming.  An interesting future direction will be to make PATH more resilient in jammed environments.  Finally, we observe that any mechanism can be used to enable coarse-grained rendezvous of the two drones so they are within the same vicinity, such as using visual landmarks, RF-based localization or other methods.  PATH is focused on the fine-grained verified handoff once the drones are already proximate to each other.


\section{Conclusion}
\label{conc}

This paper presented PATH, which is shown to be a lightweight, accurate, low cost, robust, cross-platform and real time mechanism for transferring
target tracking from one UAV to another in a verified manner.  PATH achieves a cooperative UAV handoff framework by using a handoff-local 3D representation and geometric cross-view verification to transfer target tracking between two UAVs. The sender reconstructs and transfers the target location in the local frame, while the receiver projects this information into its own view for target acquisition; the returned candidate is subsequently verified through the Mutual Agreement Handshake.

Indoor geometric evaluation over 420 measurements produced mean position errors of 0.047~m for the relative UAV position and 0.030~m for the target position. In visually ambiguous outdoor scenes, PATH achieved 96.0\% frame-level receiver-side target-acquisition accuracy, with 2.0\% false-positive and 2.0\% false-negative rates, substantially outperforming neural network-based feature matching baselines. The complete handoff procedure was also demonstrated on different UAV platforms, while the implementation maintained compact communication and lightweight onboard computation.  These results demonstrate that geometry-assisted target handoff can provide an accurate, lightweight, and platform-agnostic verification layer for cooperative UAV tracking.

%% file: ref.bib
@article{10.1145/3550485,
author = {Padhy, Ram Prasad and Sa, Pankaj Kumar and Narducci, Fabio and Bisogni, Carmen and Bakshi, Sambit},
title = {Monocular Vision-aided Depth Measurement from RGB Images for Autonomous UAV Navigation},
year = {2023},
issue_date = {February 2024},
publisher = {Association for Computing Machinery},
address = {New York, NY, USA},
volume = {20},
number = {2},
issn = {1551-6857},
url = {https://doi.org/10.1145/3550485},
doi = {10.1145/3550485},
month = sep,
articleno = {37},
numpages = {22}
}

@ARTICLE{9409130,
  author={Cavagna, Andrea and Feng, Xiao and Melillo, Stefania and Parisi, Leonardo and Postiglione, Lorena and Villegas, Pablo},
  journal={IEEE Transactions on Instrumentation and Measurement}, 
  title={CoMo: A Novel Comoving 3D Camera System}, 
  year={2021},
  volume={70},
  number={},
  pages={1-16},
  doi={10.1109/TIM.2021.3074388}}

@ARTICLE{10620329,
  author={Kubo, Nobuaki},
  journal={IEICE Transactions on Communications}, 
  title={Global Navigation Satellite System Precise Positioning Technology}, 
  year={2024},
  volume={E107-B},
  number={11},
  pages={691-705},
  doi={10.23919/transcom.2024EBI0001}}

@ARTICLE{11271135,
  author={Kim, Noah Minchan and Min, Dongchan and Nam, Gihun and Lee, Jiyun},
  journal={IEEE Transactions on Aerospace and Electronic Systems}, 
  title={Position Domain Inertial-Aided Ambiguity Resolution for High-Integrity Single-Epoch RTK}, 
  year={2026},
  volume={62},
  number={},
  pages={1426-1440},
  doi={10.1109/TAES.2025.3634315}}

@ARTICLE{visiondronedetect2025wang,
  author={Wang, Ban and Li, Jun and Zhou, Maoying and Lu, Qinfen},
  journal={IEEE Sensors Journal}, 
  title={Drone Detection and Tracking: An Edge-Deployable Efficient Algorithm Based on Vision Sensor}, 
  year={2025},
  volume={25},
  number={17},
  pages={34126-34140},
  doi={10.1109/JSEN.2025.3588414}}

@ARTICLE{10778212,
  author={Abdelkader, Mohamed and Gabr, Khaled and Jarraya, Imen and AlMusalami, Abdullah and Koubaa, Anis},
  journal={IEEE Sensors Journal}, 
  title={SMART-TRACK: A Novel Kalman Filter-Guided Sensor Fusion for Robust UAV Object Tracking in Dynamic Environments}, 
  year={2025},
  volume={25},
  number={2},
  pages={3086-3097},
  doi={10.1109/JSEN.2024.3505939}}

@ARTICLE{10552170,
  author={Yang, Ziqin and Xie, Fuxin and Zhou, Jian and Yao, Yuan and Hu, Cheng and Zhou, Baoding},
  journal={IEEE Sensors Journal}, 
  title={AIGDet: Altitude-Information-Guided Vehicle Target Detection in UAV-Based Images}, 
  year={2024},
  volume={24},
  number={14},
  pages={22672-22684},
  doi={10.1109/JSEN.2024.3406540}}

@article{mishra2020drone,
  title={Drone-surveillance for search and rescue in natural disaster},
  author={Mishra, Balmukund and Garg, Deepak and Narang, Pratik and Mishra, Vipul},
  journal={Computer Communications},
  volume={156},
  pages={1--10},
  year={2020},
  publisher={Elsevier}
}

@article{akram2024dronessl,
  title={Dronessl: Self-supervised multimodal anomaly detection in internet of drone things},
  author={Akram, Junaid and Anaissi, Ali and Othman, Wajdy and Alabdulatif, Abdulatif and Akram, Awais},
  journal={IEEE Transactions on Consumer Electronics},
  year={2024},
  publisher={IEEE}
}

@article{han2024event,
  title={Event-Assisted Object Tracking on High-Speed Drones in Harsh Illumination Environment},
  author={Han, Yuqi and Yu, Xiaohang and Luan, Heng and Suo, Jinli},
  journal={Drones},
  volume={8},
  number={1},
  pages={22},
  year={2024},
  publisher={MDPI}
}

@article{yeom2024thermal,
  title={Thermal image tracking for search and rescue missions with a drone},
  author={Yeom, Seokwon},
  journal={Drones},
  volume={8},
  number={2},
  pages={53},
  year={2024},
  publisher={MDPI}
}

@article{fu2017multi,
  title={Multi-UAVs Cooperative Localization Algorithms with Communication Constraints},
  author={Fu, Xiaowei and Bi, Haiyang and Gao, Xiaoguang},
  journal={Mathematical Problems in Engineering},
  volume={2017},
  number={1},
  pages={1943539},
  year={2017},
  publisher={Wiley Online Library}
}

@article{lee2013cooperative,
  title={Cooperative localization between small UAVs using a combination of heterogeneous sensors},
  author={Lee, Wonsuk and Bang, Hyochoong and Leeghim, Henzeh},
  journal={Aerospace science and technology},
  volume={27},
  number={1},
  pages={105--111},
  year={2013},
  publisher={Elsevier}
}

@inproceedings{cui2015drones,
  title={Drones for cooperative search and rescue in post-disaster situation},
  author={Cui, Jin Q and Phang, Swee King and Ang, Kevin ZY and Wang, Fei and Dong, Xiangxu and Ke, Yijie and Lai, Shupeng and Li, Kun and Li, Xiang and Lin, Feng and others},
  booktitle={2015 IEEE 7th international conference on cybernetics and intelligent systems (CIS) and IEEE conference on robotics, automation and mechatronics (RAM)},
  pages={167--174},
  year={2015},
  organization={IEEE}
}

@inproceedings{he2020city,
  title={City-scale multi-camera vehicle tracking by semantic attribute parsing and cross-camera tracklet matching},
  author={He, Yuhang and Han, Jie and Yu, Wentao and Hong, Xiaopeng and Wei, Xing and Gong, Yihong},
  booktitle={Proceedings of the IEEE/CVF Conference on Computer Vision and Pattern Recognition Workshops},
  pages={576--577},
  year={2020}
}

@inproceedings{specker2022improving,
  title={Improving multi-target multi-camera tracking by track refinement and completion},
  author={Specker, Andreas and Florin, Lucas and Cormier, Mickael and Beyerer, J{\"u}rgen},
  booktitle={Proceedings of the IEEE/CVF Conference on Computer Vision and Pattern Recognition},
  pages={3199--3209},
  year={2022}
}

@inproceedings{yang2022box,
  title={Box-grained reranking matching for multi-camera multi-target tracking},
  author={Yang, Xipeng and Ye, Jin and Lu, Jincheng and Gong, Chenting and Jiang, Minyue and Lin, Xiangru and Zhang, Wei and Tan, Xiao and Li, Yingying and Ye, Xiaoqing and others},
  booktitle={Proceedings of the IEEE/CVF conference on computer vision and pattern recognition},
  pages={3096--3106},
  year={2022}
}

@inproceedings{liu2021city,
  title={City-scale multi-camera vehicle tracking guided by crossroad zones},
  author={Liu, Chong and Zhang, Yuqi and Luo, Hao and Tang, Jiasheng and Chen, Weihua and Xu, Xianzhe and Wang, Fan and Li, Hao and Shen, Yi-Dong},
  booktitle={Proceedings of the IEEE/CVF Conference on Computer Vision and Pattern Recognition},
  pages={4129--4137},
  year={2021}
}

@article{hsu2021multi,
  title={Multi-target multi-camera tracking of vehicles using metadata-aided re-id and trajectory-based camera link model},
  author={Hsu, Hung-Min and Cai, Jiarui and Wang, Yizhou and Hwang, Jenq-Neng and Kim, Kwang-Ju},
  journal={IEEE Transactions on Image Processing},
  volume={30},
  pages={5198--5210},
  year={2021},
  publisher={IEEE}
}

@article{fang2024strategies,
  title={Strategies for optimized uav surveillance in various tasks and scenarios: A review},
  author={Fang, Zixuan and Savkin, Andrey V},
  journal={Drones},
  volume={8},
  number={5},
  pages={193},
  year={2024},
  publisher={MDPI}
}

@article{thakur2021artificial,
  title={Artificial intelligence techniques in smart cities surveillance using UAVs: A survey},
  author={Thakur, Narina and Nagrath, Preeti and Jain, Rachna and Saini, Dharmender and Sharma, Nitika and Hemanth, D Jude},
  journal={Machine Intelligence and Data Analytics for Sustainable Future Smart Cities},
  pages={329--353},
  year={2021},
  publisher={Springer}
}

@article{alsamhi2022uav,
  title={UAV computing-assisted search and rescue mission framework for disaster and harsh environment mitigation},
  author={Alsamhi, Saeed Hamood and Shvetsov, Alexey V and Kumar, Santosh and Shvetsova, Svetlana V and Alhartomi, Mohammed A and Hawbani, Ammar and Rajput, Navin Singh and Srivastava, Sumit and Saif, Abdu and Nyangaresi, Vincent Omollo},
  journal={Drones},
  volume={6},
  number={7},
  pages={154},
  year={2022},
  publisher={MDPI}
}

@article{dahal2021design,
  title={Design and Analysis of Propeller for High-Altitude Search and Rescue Unmanned Aerial Vehicle},
  author={Dahal, Chiranjivi and Dura, Hari Bahadur and Poudel, Laxman},
  journal={International Journal of Aerospace Engineering},
  volume={2021},
  number={1},
  pages={6629489},
  year={2021},
  publisher={Wiley Online Library}
}

@article{lun2022target,
  title={Target search in dynamic environments with multiple solar-powered UAVs},
  author={Lun, Yuebin and Wang, Honglun and Wu, Jianfa and Liu, Yiheng and Wang, Yanxiang},
  journal={IEEE Transactions on Vehicular Technology},
  volume={71},
  number={9},
  pages={9309--9321},
  year={2022},
  publisher={IEEE}
}

@article{ASADZADEH2022109633,
title = {UAV-based remote sensing for the petroleum industry and environmental monitoring: State-of-the-art and perspectives},
journal = {Journal of Petroleum Science and Engineering},
volume = {208},
pages = {109633},
year = {2022},
issn = {0920-4105},
author = {Saeid Asadzadeh and Wilson José de Oliveira and Carlos Roberto de {Souza Filho}}
}

@article{kuantama2024laser,
  title={Laser-based drone vision disruption with a real-time tracking system for privacy preservation},
  author={Kuantama, Endrowednes and Zhang, Yihao and Rahman, Faiyaz and Han, Richard and Dawes, Judith and Mildren, Rich and Abir, Tasnim Azad and Nguyen, Phuc},
  journal={Expert Systems with Applications},
  volume={255},
  pages={124626},
  year={2024},
  publisher={Elsevier}
}

@article{elbers2021efficacy,
  title={Efficacy of an automated laser for reducing wild bird visits to the free range area of a poultry farm},
  author={Elbers, Armin RW and Gonzales, Jos{\'e} L},
  journal={Scientific Reports},
  volume={11},
  number={1},
  pages={12779},
  year={2021},
  publisher={Nature Publishing Group UK London}
}

@article{eisenbeiser2022gills,
  title={Gills just want to have fun: Can fish play games, just like us?},
  author={Eisenbeiser, Sofia and Serbe-Kamp, {\'E}tienne and Gage, Gregory J and Marzullo, Timothy C},
  journal={Animals},
  volume={12},
  number={13},
  pages={1684},
  year={2022},
  publisher={Multidisciplinary Digital Publishing Institute}
}

@article{grigg2024associations,
  title={Associations between Laser Light Pointer Play and Repetitive Behaviors in Companion Cats: Does Participant Recruitment Method Matter?},
  author={Grigg, Emma K and Kogan, Lori R},
  journal={Journal of Applied Animal Welfare Science},
  volume={27},
  number={2},
  pages={250--265},
  year={2024},
  publisher={Taylor \& Francis}
}

@INPROCEEDINGS{803809,
  author={Kato, H. and Billinghurst, M.},
  booktitle={Proceedings 2nd IEEE and ACM International Workshop on Augmented Reality (IWAR'99)}, 
  title={Marker tracking and HMD calibration for a video-based augmented reality conferencing system}, 
  year={1999},
  volume={},
  number={},
  pages={85-94}
}

@inproceedings{LiangSrigrarom_ICUAS2021,
  author    = {Liang, Niven Sie Jun and Srigrarom, Sutthiphong},
  title     = {Multi-camera multi-target drone tracking systems with trajectory-based target matching and re-identification},
  booktitle = {2021 International Conference on Unmanned Aircraft Systems (ICUAS)},
  year      = {2021},
  pages     = {1337--1344},
  address   = {Athens, Greece},
  month     = {Jun},
  publisher = {IEEE},
  isbn      = {978-1-6654-1535-4}
}

@article{Zhan_Chen_Chen_Zhang_2025,
  author    = {Zhan, Xinru and Chen, Yang and Chen, Xi and Zhang, Wenhao},
  title     = {Balanced multi-{UAV} path planning for persistent monitoring},
  journal   = {Robotica},
  year      = {2025},
  volume    = {43},
  number    = {1},
  pages     = {332--349},
  month     = jan,
  publisher = {Cambridge University Press},
  note      = {First published online: 20 Nov 2024}
}

@inproceedings{bauer2024persistent,
  author    = {Bauer, Maximilian and Alhamwy, Yasin and Geihs, Kurt},
  title     = {Persistent UAV Formation Flight by Dynamic Agent Replacement and Leader Selection},
  booktitle = {Progressive and Integrative Ideas and Applications of Engineering Systems Under the Framework of IOT and AI},
  editor    = {Ma, Yongsheng},
  series    = {Lecture Notes in Electrical Engineering},
  volume    = {1076},
  pages     = {106--117},
  year      = {2024},
  publisher = {Springer},
  address   = {Singapore},
  note      = {ISDEA 2023, Okayama, Japan, 12--14 May 2023}
}

@article{Upadhyay_Rawat_Deb_2021,
  author       = {Upadhyay, Jatin and Rawat, Abhishek and Deb, Dipankar},
  title        = {Multiple Drone Navigation and Formation Using Selective Target Tracking-Based Computer Vision},
  journal      = {Electronics},
  year         = {2021},
  volume       = {10},
  number       = {17},
  articlenumber= {2125},
  month        = sep,
  publisher    = {MDPI}
}

@article{tavasci2024reliability,
  title={Reliability of real-time kinematic (RTK) positioning for low-cost drones’ navigation across global navigation satellite System (GNSS) critical environments},
  author={Tavasci, Luca and Nex, Francesco and Gandolfi, Stefano},
  journal={Sensors (Basel, Switzerland)},
  volume={24},
  number={18},
  pages={6096},
  year={2024}
}

@article{ming2022laser,
  author    = {Ming, Rui and Zhou, Zhiyan and Lyu, Zichen and Luo, Xiwen and Zi, Le and Song, Cancan and Zang, Yu and Liu, Wei and Jiang, Rui},
  title     = {Laser tracking leader-follower automatic cooperative navigation system for UAVs},
  journal   = {International Journal of Agricultural and Biological Engineering},
  year      = {2022},
  volume    = {15},
  number    = {2},
  pages     = {165--176},
  publisher = {Chinese Society of Agricultural Engineering},
  issn      = {1934-6344}
}

@article{pengnianmultitrack2024_drones,
  author  = {Wu, Pengnian and Li, Yixuan and Xue, Dong},
  title   = {Multi-Target Tracking with Multiple Unmanned Aerial Vehicles Based on Information Fusion},
  journal = {Drones},
  year    = {2024},
  volume  = {8},
  number  = {12},
  pages   = {704}
}

@inproceedings{kim2025continuous,
  title={Continuous Marine Monitoring via Autonomous UAV Handoff},
  author={Kim, Heegyeong and James, Alice and Seth, Avishkar and Kuantama, Endrowednes and Williamson, Jane and Feng, Yimeng and Han, Richard},
  booktitle={Proceedings of the 23rd Annual International Conference on mobile systems, applications and services},
  pages={711--716},
  year={2025}
}

@article{haalck2023cater,
  title={CATER: combined animal tracking \& environment reconstruction},
  author={Haalck, Lars and Mangan, Michael and Wystrach, Antoine and Clement, Leo and Webb, Barbara and Risse, Benjamin},
  journal={Science Advances},
  volume={9},
  number={16},
  pages={eadg2094},
  year={2023},
  publisher={American Association for the Advancement of Science}
}

@article{zhu2022visdrone,
  author  = {Zhu, Pengfei and Wen, Longyin and Du, Dawei and Bian, Xiao and Fan, Heng and Hu, Qinghua and Ling, Haibin},
  title   = {Detection and Tracking Meet Drones Challenge},
  journal = {IEEE Transactions on Pattern Analysis and Machine Intelligence},
  year    = {2022},
  volume  = {44},
  number  = {11},
  pages   = {7380--7399},
}

@article{ye2022survey,
  author  = {Ye, Mang and Shen, Jianbing and Lin, Gaojie and Xiang, Tao and Shao, Ling and Hoi, Steven C. H.},
  title   = {Deep Learning for Person Re-Identification: A Survey and Outlook},
  journal = {IEEE Transactions on Pattern Analysis and Machine Intelligence},
  year    = {2022},
  volume  = {44},
  number  = {6},
  pages   = {2872--2893},
}

@inproceedings{rublee2011orb,
  title={ORB: An efficient alternative to SIFT or SURF},
  author={Rublee, Ethan and Rabaud, Vincent and Konolige, Kurt and Bradski, Gary},
  booktitle={2011 International conference on computer vision},
  pages={2564--2571},
  year={2011},
  organization={Ieee}
}

@INPROCEEDINGS{potje2024cvpr,
  author={Potje, Guilherme and Cadar, Felipe and Araujo, André and Martins, Renato and Nascimento, Erickson R.},
  booktitle={2024 IEEE/CVF Conference on Computer Vision and Pattern Recognition (CVPR)}, 
  title={XFeat: Accelerated Features for Lightweight Image Matching}, 
  year={2024},
  pages={2682-2691},
  doi={10.1109/CVPR52733.2024.00259}}

@inproceedings{lourencco2021intel,
  title={Intel RealSense SR305, D415 and L515: Experimental Evaluation and Comparison of Depth Estimation.},
  author={Louren{\c{c}}o, Francisco and Araujo, Helder},
  booktitle={VISIGRAPP (4: VISAPP)},
  pages={362--369},
  year={2021}
}

@article{pfleging2015dynamic,
  title={Dynamic monitoring reveals motor task characteristics in prehistoric technical gestures},
  author={Pfleging, Johannes and St{\"u}cheli, Marius and Iovita, Radu and Buchli, Jonas},
  journal={PloS one},
  volume={10},
  number={8},
  pages={e0134570},
  year={2015},
  publisher={Public Library of Science San Francisco, CA USA}
}

@article{servi2024comparative,
  title={Comparative evaluation of Intel RealSense D415, D435i, D455, and Microsoft azure kinect DK sensors for 3D vision applications},
  author={Servi, Michaela and Profili, Andrea and Furferi, Rocco and Volpe, Yary},
  journal={IEEE Access},
  volume={12},
  pages={111311--111321},
  year={2024},
  publisher={IEEE}
}

@manual{intelrealsense2019,
  title        = {Intel® RealSense™ D400 Series Product Family Datasheet},
  author       = {{Intel Corporation}},
  year         = {2019},
  month        = {October},
  note         = {Rev. 005},
  url          = 
{https://www.realsenseai.com/wp-content/uploads/2019/10/Intel-RealSense-D400-Series-Datasheet-Oct-2019.pdf}
}
